\newif\ifejs
\ejsfalse

\ifejs
	\documentclass[ejsv2]{imsart}

	\RequirePackage[authoryear]{natbib}
	\RequirePackage[colorlinks,citecolor=blue,urlcolor=blue]{hyperref}
	\RequirePackage{graphicx}

	\arxiv{2502.13570}
	\startlocaldefs
	\endlocaldefs

	\usepackage[dvipsnames]{xcolor}
	\usepackage[poorman]{cleveref}

\else
	\documentclass[a4paper]{article}
	\usepackage[a4paper, total={15.7cm, 22cm}]{geometry}
	\usepackage[utf8]{inputenc}
	\usepackage[pdftex,colorlinks=true,linkcolor=blue,citecolor=magenta]{hyperref} 
	\usepackage[dvipsnames]{xcolor}

	\usepackage{amsmath}
	\usepackage{amssymb}
	\usepackage{mathtools}
	\usepackage{csquotes}
	\usepackage{amsfonts}

	\usepackage{cleveref}
\fi

\usepackage{dsfont}
\usepackage[]{subfigure}
\usepackage{xparse}

\usepackage[ruled,algosection,vlined,algo2e]{algorithm2e} 

\crefalias{algocfline}{line}
\crefname{algocf}{algorithm}{algorithms}
\Crefname{algocf}{Algorithm}{Algorithms}

\usepackage{enumerate}
\usepackage[overload]{empheq}
\usepackage{nicefrac}
\usepackage[normalem]{ulem}
\usepackage{xargs}                      
\definecolor{bluee}{HTML}{4967ed}
\definecolor{malgared}{rgb}{0.8627450980392157, 0.34901960784313724, 0.34901960784313724}

\usepackage{booktabs,boldline}
\usepackage{multirow}
\usepackage{makecell}

\usepackage{enumitem}
\setitemize{noitemsep,topsep=2pt,parsep=2pt,partopsep=0pt}
\setenumerate{noitemsep,topsep=2pt,parsep=2pt,partopsep=0pt}

\DeclareDocumentCommand\new{g}{#1}

\usepackage{import}
\usepackage{xspace}
\usepackage{braket} 	
\usepackage{commath} 

\newcommand{\bN}{\mathbb{N}}

\newcommand{\bR}{\mathbb{R}}

\usepackage{pgffor}
\foreach \x in {a,...,z}{
  \expandafter\xdef\csname V\x \endcsname{\noexpand\ensuremath{\noexpand\V{\x}}}
}
\foreach \x in {A,...,Z}{
  \expandafter\xdef\csname V\x \endcsname{\noexpand\ensuremath{\noexpand\V{\x}}}
  \expandafter\xdef\csname c\x \endcsname{\noexpand\ensuremath{\noexpand\mathcal{\x}}}
  \expandafter\xdef\csname f\x \endcsname{\noexpand\ensuremath{\noexpand\mathfrak{\x}}}
}

\DeclareMathOperator*{\rg}{ran}

\DeclareMathOperator*{\rk}{rk}
\DeclareMathOperator*{\esssup}{ess\,sup}

\DeclareMathOperator*{\Tr}{tr}
\DeclareMathOperator*{\spa}{span}
\ifLuaTeX
	\newcommand{\V}[1]{\symbf{#1}} 
\else
	\usepackage{bm}
	\newcommand{\V}[1]{\bm{#1}} 
\fi

\newcommand\de{:=}
\newcommand\eg{e.g.\xspace }
\newcommand\ie{i.e.\xspace }
\newcommand\wrt{w.r.t.\xspace }

\NewDocumentCommand\irange{O{1}m}{\brk*{#1,…,#2}}
\newcommand{\diid}{\overset{{\small{i.i.d.}}}{\sim}}

\newcommand{\tiid}{i.i.d.\ }

\newcommand\restr[2]{{
  \left.\kern-\nulldelimiterspace 
  #1 
  \vphantom{\big|} 
  \right|_{#2} 
  }}

\newcommand{\kron}{\otimes}

\newcommand{\vvec}[1]{\begin{bmatrix}#1\end{bmatrix}}

\newcommand{\E}{\mathbf{E}}

\newcommand{\MMD}{\textup{MMD}}

\providecommand\given{}
\DeclarePairedDelimiterX{\brkg}[1]{[}{]}{
\renewcommand\given{\nonscript\:\delimsize\vert\nonscript\:\mathopen{}}%
#1}
\RenewDocumentCommand{\P}{se{_^}sd[]}{%
	\mathbb{P}\IfValueT{#2}{_{#2}}\IfValueT{#3}{^{#3}}%
	\IfBooleanTF{#1}{%
		\IfValueT{#5}{\brkg*{#5}}%
	}{%
		\IfBooleanTF{#4}{%
			\IfValueT{#5}{\brkg*{#5}}%
		}{%
			\IfValueT{#5}{\brkg{#5}}%
		}%
	}
}

\DeclarePairedDelimiter{\prt}{(}{)}
\DeclarePairedDelimiter{\brk}{[}{]}
\DeclarePairedDelimiter{\cb}{\{}{\}}
\let\norm\relax
\DeclarePairedDelimiter{\norm}{\lVert}{\rVert}
\DeclarePairedDelimiter{\n}{\lVert}{\rVert}
\DeclarePairedDelimiter{\ip}{\langle}{\rangle}

\DeclarePairedDelimiter{\absv}{|}{|}

\DeclareFontFamily{U}{matha}{\hyphenchar\font45}
\DeclareFontShape{U}{matha}{m}{n}{
<-6> matha5 <6-7> matha6 <7-8> matha7
<8-9> matha8 <9-10> matha9
<10-12> matha10 <12-> matha12
}{}
\DeclareSymbolFont{matha}{U}{matha}{m}{n}

\DeclareFontFamily{U}{mathx}{\hyphenchar\font45}
\DeclareFontShape{U}{mathx}{m}{n}{
<-6> mathx5 <6-7> mathx6 <7-8> mathx7
<8-9> mathx8 <9-10> mathx9
<10-12> mathx10 <12-> mathx12
}{}
\DeclareSymbolFont{mathx}{U}{mathx}{m}{n}

\DeclareMathDelimiter{\vvvert} {0}{matha}{"7E}{mathx}{"17}%
\DeclarePairedDelimiterX{\normiii}[1]
{\vvvert}
{\vvvert}
{\ifblank{#1}{\:\cdot\:}{#1}}

\usepackage{pifont}

\newcommand\mdeff[1][λ]{d_{\textup{eff}}\prt*{#1}}
\newcommand\emdeff[1][λ]{\hat{d}_{\textup{eff}}\prt*{#1}}
\newcommand\dsp{\cZ} 	

\DeclareDocumentCommand\als{g}{\hat{\ell}_{λ}\IfNoValueF{#1}{(#1)}} 
\DeclareDocumentCommand\tls{g}{\ell_{λ}\IfNoValueF{#1}{(#1)}} 
\NewDocumentCommand\kmat{O{n}}{K_{#1}}
\DeclareDocumentCommand\pals{g}{p_λ\IfNoValueF{#1}{(#1)}} 

\newcommand\rkhs{\cH} 	
\newcommand\supfmap{‖\kr‖_∞^{1/2}}
\newcommand\supk{‖\kr‖_∞}

\DeclarePairedDelimiter{\nrkhs}{\lVert}{\rVert_{\new{\rkhs}}}	
\DeclarePairedDelimiter{\noprkhs}{\lVert}{\rVert}	
\DeclarePairedDelimiter{\iprkhs}{\langle}{\rangle_{\new{\rkhs}}}	

\newcommand\at{T_{\nX,\nY}} 	

\newcommand\tsl{\hat{Ψ}}
\NewDocumentCommand{\ts}{e{^}O{X,Y}}{\tsl\IfValueT{#1}{^{#1}}\IfNoValueF{#2}{(#2)}}
\NewDocumentCommand\pts{O{σ}}{\ts^{#1}}
\NewDocumentCommand\pstsU{O{X,Y}}{\hat{U}^σ\IfNoValueF{#1}{(#1)}}
\NewDocumentCommand\pstsR{O{X,Y}}{\hat{R}^σ\IfNoValueF{#1}{(#1)}}
\NewDocumentCommand\nysts{O{X,Y}}{\hat{Ψ}_{\textup{Nys}}\IfNoValueF{#1}{(#1)}}
\NewDocumentCommand\ipts{g}{\hat{Ψ}\IfNoValueF{#1}{_{#1}}} 
\NewDocumentCommand\oipts{g}{\hat{Ψ}\IfValueT{#1}{_{(#1)}}} 

\newcommand\X{X}
\newcommand\Y{Y}
\newcommand\XY{W}
\newcommand\W\XY
\newcommand\nX{n_X}
\newcommand\nY{n_Y}
\newcommand\dX{\hat{P}}
\newcommand\dY{\hat{Q}}

\newcommand\covP{C_P}
\newcommand\covQ{C_Q}
\newcommand\covM{\bar{C}}	
\newcommand\rcovM{\bar{C}_{λ}}	
\newcommand\ecovX{C_{\X}}
\newcommand\ecovY{C_{\Y}}
\newcommand\ecovM{\bar{C}_{n}}	
\newcommand\recovM{\bar{C}_{n,λ}}	

\newcommand\nf{\ell}	

\newcommand\np{\cP} 
\NewDocumentCommand\kme{g}{μ\IfValueT{#1}{{(#1)}}} 
\NewDocumentCommand\kmeX{}{μ(\X)} 
\NewDocumentCommand\kmeY{}{μ(\Y)} 
\NewDocumentCommand\bkme{m}{μ\IfValueT{#1}{_{#1}}} 
\NewDocumentCommand\bakme{m}{\tfmap\IfValueT{#1}{_{#1}}} 

\newcommand\tq[1][α]{q_{1-#1}(\X,\Y)}	
\newcommand\pstsUq{q_{1-α,\hat{U}^σ}(\X,\Y)}	

\newcommand\thr{\oipts{b_α}}

\newcommand\ldms{Z}
\DeclareDocumentCommand\ldm{g}{z\IfNoValueF{#1}{_{#1}}}
\DeclareDocumentCommand\ldmX{g}{\tilde{x}\IfNoValueF{#1}{_{#1}}}
\DeclareDocumentCommand\ldmY{g}{\tilde{y}\IfNoValueF{#1}{_{#1}}}

\DeclareDocumentCommand\ftldms{}{Φ_{Z}}
\DeclareDocumentCommand\ftn{}{Φ_W}

\DeclareDocumentCommand\eMMDv{}{\cE_{\MMD}}
\DeclareDocumentCommand\eMMD{O{δ}O{\nX}O{\nY}}{\cE_{\MMD}(#2,#3,#1)}
\DeclareDocumentCommand\eKME{O{δ}O{n}}{\cE_{\textup{KME}}(#2,#1)}
\DeclareDocumentCommand\eb{O{δ}O{n}}{\cE_{b}(#2,#1)}	
\DeclareDocumentCommand\edb{O{δ}}{\cE_{b}(\nX,\nY,#1)}	

\newcommand\Kl{K_{\ldms}}
\newcommand\Hm{\cH_{\nf}}
\newcommand\Pm{P_{\ldms}}
\newcommand\Pmo{\Pm^⟂}
\newcommand\Pl{\Pm}

\NewDocumentCommand\kr{g}{κ\IfValueT{#1}{{(#1)}}}		
\NewDocumentCommand\fmap{g}{ϕ\IfValueT{#1}{{(#1)}}}				

\NewDocumentCommand\tk{g}{\tilde{κ}\IfValueT{#1}{{(#1)}}}		
\NewDocumentCommand\tfmap{g}{φ\IfValueT{#1}{{(#1)}}}		

\DeclareUnicodeCharacter{03B1}{\alpha}  	 
\DeclareUnicodeCharacter{0391}{\Alpha}       
\DeclareUnicodeCharacter{03B2}{\beta}        
\DeclareUnicodeCharacter{0392}{\Beta}        
\DeclareUnicodeCharacter{03B3}{\gamma}       
\DeclareUnicodeCharacter{0393}{\Gamma}       
\DeclareUnicodeCharacter{03B4}{\delta}       
\DeclareUnicodeCharacter{0394}{\Delta}       
\DeclareUnicodeCharacter{03B5}{\epsilon}     
\DeclareUnicodeCharacter{0395}{\Epsilon}     
\DeclareUnicodeCharacter{03B6}{\zeta}        
\DeclareUnicodeCharacter{0396}{\Zeta}        
\DeclareUnicodeCharacter{03B7}{\eta}         
\DeclareUnicodeCharacter{0397}{\Eta}         
\DeclareUnicodeCharacter{03B8}{\theta}       
\DeclareUnicodeCharacter{0398}{\Theta}       
\DeclareUnicodeCharacter{03B9}{\iota}        
\DeclareUnicodeCharacter{0399}{\Iota}        
\DeclareUnicodeCharacter{03BA}{\kappa}       
\DeclareUnicodeCharacter{039A}{\Kappa}       
\DeclareUnicodeCharacter{03BB}{\lambda}      
\DeclareUnicodeCharacter{039B}{\Lambda}      
\DeclareUnicodeCharacter{03BC}{\mu}          
\DeclareUnicodeCharacter{039C}{\Mu}          
\DeclareUnicodeCharacter{03BD}{\nu}          
\DeclareUnicodeCharacter{039D}{\Nu}          
\DeclareUnicodeCharacter{03BE}{\xi}          
\DeclareUnicodeCharacter{039E}{\Xi}          
\DeclareUnicodeCharacter{03BF}{\omega}       
\DeclareUnicodeCharacter{039F}{\Omega}       
\DeclareUnicodeCharacter{03C0}{\pi}          
\DeclareUnicodeCharacter{03A0}{\Pi}          
\DeclareUnicodeCharacter{03C1}{\rho}         
\DeclareUnicodeCharacter{03A1}{\Rho}         
\DeclareUnicodeCharacter{03C3}{\sigma}       
\DeclareUnicodeCharacter{03A3}{\Sigma}       
\DeclareUnicodeCharacter{03C4}{\tau}         
\DeclareUnicodeCharacter{03A4}{\Tau}         
\DeclareUnicodeCharacter{03C5}{\upsilon}     
\DeclareUnicodeCharacter{03A5}{\Upsilon}     
\DeclareUnicodeCharacter{03D5}{\phi}         
\DeclareUnicodeCharacter{03C6}{\varphi}      
\DeclareUnicodeCharacter{03A6}{\Phi}         
\DeclareUnicodeCharacter{03C7}{\xi}          
\DeclareUnicodeCharacter{03A7}{\Xi}          
\DeclareUnicodeCharacter{03C8}{\psi}         
\DeclareUnicodeCharacter{03A8}{\Psi}         
\DeclareUnicodeCharacter{03C9}{\omega}       
\DeclareUnicodeCharacter{03A9}{\Omega}       
\DeclareUnicodeCharacter{2113}{\ell}      	 
\DeclareUnicodeCharacter{1D62}{_i}       	 %

\DeclareUnicodeCharacter{2248}{\approx}      
\DeclareUnicodeCharacter{2264}{\leq}         
\DeclareUnicodeCharacter{2265}{\geq}         
\DeclareUnicodeCharacter{227C}{\preccurlyeq} 
\DeclareUnicodeCharacter{227D}{\succcurlyeq} 
\DeclareUnicodeCharacter{2286}{\subseteq}    
\DeclareUnicodeCharacter{2260}{\neq}   		 
\DeclareUnicodeCharacter{1D40}{\ifmmode^T\else\textsuperscript{T}\fi}
\DeclareUnicodeCharacter{2200}{\forall}      
\DeclareUnicodeCharacter{2203}{\exists}      
\DeclareUnicodeCharacter{2207}{\nabla}       
\DeclareUnicodeCharacter{220A}{\in}          
\DeclareUnicodeCharacter{00B7}{\cdot}        
\DeclareUnicodeCharacter{00D7}{\times}       
\DeclareUnicodeCharacter{00B7}{\cdot}        
\DeclareUnicodeCharacter{2016}{\Vert}        
\DeclareUnicodeCharacter{2026}{\dots}        
\DeclareUnicodeCharacter{2020}{\dagger}      
\DeclareUnicodeCharacter{27C2}{\perp}        %
\DeclareUnicodeCharacter{2229}{\cap}        
\DeclareUnicodeCharacter{222A}{\cup}        
\DeclareUnicodeCharacter{00BD}{\nicefrac{1}{2}} 
\DeclareUnicodeCharacter{2190}{\leftarrow}   
\DeclareUnicodeCharacter{2192}{\rightarrow}  
\DeclareUnicodeCharacter{2194}{\leftrightarrow} %
\DeclareUnicodeCharacter{21A6}{\mapsto}       
\DeclareUnicodeCharacter{21D2}{\implies} 	 
\DeclareUnicodeCharacter{21D4}{\iff} 		 
\DeclareUnicodeCharacter{211D}{\mathbb{R}}   
\DeclareUnicodeCharacter{2102}{\mathbb{C}}   
\DeclareUnicodeCharacter{221E}{\infty}   
\DeclareUnicodeCharacter{00BD}{\nicefrac{1}{2}} 
\DeclareUnicodeCharacter{00B1}{\pm}            
\DeclareUnicodeCharacter{00B9}{\ifmmode^1\else\textsuperscript{1}\fi}           
\DeclareUnicodeCharacter{00B2}{\ifmmode^2\else\textsuperscript{2}\fi}           
\DeclareUnicodeCharacter{00B3}{\ifmmode^3\else\textsuperscript{3}\fi}           
\DeclareUnicodeCharacter{2074}{\ifmmode^4\else\textsuperscript{4}\fi}           
\DeclareUnicodeCharacter{2075}{\ifmmode^5\else\textsuperscript{5}\fi}           
\DeclareUnicodeCharacter{2076}{\ifmmode^6\else\textsuperscript{6}\fi}           
\DeclareUnicodeCharacter{2077}{\ifmmode^7\else\textsuperscript{7}\fi}           
\DeclareUnicodeCharacter{2078}{\ifmmode^8\else\textsuperscript{8}\fi}           
\DeclareUnicodeCharacter{2079}{\ifmmode^9\else\textsuperscript{9}\fi}           
\DeclareUnicodeCharacter{1D62}{_i}           %
\DeclareUnicodeCharacter{2C7C}{_j}           %
\DeclareUnicodeCharacter{2096}{_k}           %
\DeclareUnicodeCharacter{2097}{_l}           %
\DeclareUnicodeCharacter{2080}{_0}           
\DeclareUnicodeCharacter{2081}{_1}           
\DeclareUnicodeCharacter{2082}{_2}           
\DeclareUnicodeCharacter{2083}{_3}           
\DeclareUnicodeCharacter{2084}{_4}           
\DeclareUnicodeCharacter{2085}{_5}           
\DeclareUnicodeCharacter{2086}{_6}           
\DeclareUnicodeCharacter{2087}{_7}           
\DeclareUnicodeCharacter{2088}{_8}           
\DeclareUnicodeCharacter{2089}{_9}           

\DeclareUnicodeCharacter{2212}{-}            
\DeclareUnicodeCharacter{207B}{\ifmmode^{-}\else\textsuperscript{-}\fi}            
\DeclareUnicodeCharacter{2013}{--}           
\DeclareUnicodeCharacter{2014}{---}          
\DeclareUnicodeCharacter{202F}{~}            
\DeclareUnicodeCharacter{00A0}{~}            

\ifejs
\newtheorem{definition}{Definition}
\newtheorem{lemma}{Lemma}
\newtheorem{assumption}{Assumption}
\newtheorem{theorem}{Theorem}
\crefname{definition}{definition}{definitions}
\Crefname{definition}{Definition}{Definitions}
\crefname{assumption}{assumption}{assumptions}
\Crefname{assumption}{Assumption}{Assumptions}
\crefname{algocf}{algorithm}{algorithms}
\Crefname{algocf}{Algorithm}{Algorithms}

\NewDocumentEnvironment{tlemma}{omm}{	
\begin{lemma}[#2]\label{r:#3}
}{
\end{lemma}
}
\NewDocumentEnvironment{ttheorem}{omm}{
\begin{theorem}[#2]\label{r:#3}
}{
\end{theorem}
}
\NewDocumentEnvironment{tproposition}{omm}{	
\begin{proposition}[#2]\label{r:#3}
}{
\end{proposition}
}
\NewDocumentEnvironment{tcorollary}{omm}{
\begin{corollary}[#2]\label{r:#3}
}{
\end{corollary}
}
\NewDocumentEnvironment{tdefinition}{omm}{	
\begin{definition}[#2]\label{d:#3}
}{
\end{definition}
}
\NewDocumentEnvironment{tremark}{omm}{	
\begin{remark}[#2]\label{rm:#3}
}{
\end{remark}
}
\NewDocumentEnvironment{texample}{omm}{	
\begin{example}[#2]\label{ex:#3}
}{
\end{example}
}
\NewDocumentEnvironment{tassumption}{omm}{	
\begin{assumption}[#2]\label{a:#3}
}{
\end{assumption}
}

\newenvironment{tproofof*}[1]{
\begin{proof}
}{
\end{proof}
}

\newcommand\proofparagraph[1]{\item\paragraph{#1}}
\else
\newif\iffancyversion
\fancyversiontrue
\ifejs\fancyversionfalse\fi

\usepackage{etoolbox}
\usepackage[most]{tcolorbox}
\usepackage{xstring} 

\definecolor{customBlue}{RGB}{18,75,126}
\colorlet{titleCol}{customBlue}	
\colorlet{titleThmCol}{titleCol} 
\colorlet{backCol}{titleCol!08!white}
\colorlet{backThmCol}{titleThmCol!08!white}

\tcbset{ 
  compactdefaultstyle/.style={
    use color stack, 
	breakable,
	theorem style=plain,
	enhanced,
	sharp corners,
	frame hidden,
	colback=backCol,
	opacityfill=0, 
	fontupper=\itshape,
	fonttitle=\bfseries\upshape,
	coltitle=black, 
	boxrule=0pt,
	before skip=1mm,
	after skip=1mm,
	right = 0mm, 
	left = -1mm, 
	top = 0mm,
	bottom = 0mm,
  },
  defaultthmstyle/.style={
    use color stack, 
	breakable,
	theorem style=plain,
	enhanced,
	sharp corners,
	frame hidden,
	colback=backCol,
	opacityfill=0, 
	fontupper=\itshape,
	fonttitle=\bfseries\upshape,
	coltitle=black, 
	boxrule=0pt,
	before skip=2mm,
	after skip=2mm,
	right = 0mm, 
	left = -1mm, 
	top = 1mm,
	bottom = 1mm,
  },
  defaultproofstyle/.style={
    use color stack, 
	breakable,
	theorem style=plain,
	terminator sign={.},  
	enhanced,
	sharp corners,
	frame hidden,
	colback=backCol,
	opacityfill=0, 
	fontupper=\upshape,
	fonttitle=\itshape,
	separator sign={:\ }, 
	coltitle=black, 
	boxrule=0pt,
	before skip=2mm,
	after skip=2mm,
	right = 0mm, 
	left = -1mm, 
	top = 1mm,
	bottom = 1mm,
  },
  proofstyle/.style={
    use color stack, 
	breakable,
	enhanced,
	theorem style=plain,
	frame hidden,
	sharp corners,
	colback=backCol,
	boxrule=0pt,
	before skip=2mm,
	after skip=2mm,
	right = 2mm, 
	borderline west={0.4mm}{0pt}{titleCol},
	fonttitle=\bfseries,
	coltitle=titleCol,
  }, 
  bluewestlinestyle/.style={
    use color stack, 
	breakable,
	theorem style=plain,
	enhanced,
	sharp corners,
	frame hidden,
	colback=black!03!white,
	coltitle=customBlue,
	boxrule=0pt,
	borderline west={0.8mm}{0pt}{customBlue},
	fonttitle=\upshape\bfseries\hypersetup{citecolor=customBlue,linkcolor=customBlue},
	before skip=2mm,
	after skip=2mm,
	left = 3mm, 
	right = 2mm, 
  },
  redvioletwestlinestyle/.style={
    use color stack, 
	breakable,
	theorem style=plain,
	enhanced,
	sharp corners,
	frame hidden,
	colback=black!03!white,
	coltitle=RedViolet,
	boxrule=0pt,
	borderline west={0.8mm}{0pt}{RedViolet},
	fonttitle=\upshape\bfseries\hypersetup{citecolor=RedViolet,linkcolor=RedViolet},
	before skip=2mm,
	after skip=2mm,
	left = 3mm, 
	right = 2mm, 
  },
  greenwestlinestyle/.style={
    use color stack, 
	breakable,
	theorem style=plain,
	enhanced,
	sharp corners,
	frame hidden,
	colback=black!03!white,
	coltitle=ForestGreen,
	boxrule=0pt,
	borderline west={0.8mm}{0pt}{ForestGreen},
	fonttitle=\upshape\bfseries\hypersetup{citecolor=ForestGreen,linkcolor=ForestGreen},
	before skip=2mm,
	after skip=2mm,
	left = 3mm, 
	right = 2mm, 
  },
}

\tcbset{todobox/.style={
  title={Some work is needed here}, 
  colframe=red!70!black,
  colback=red!10,
  fonttitle=\bfseries, 
  coltext=red!70!black, 
  breakable,
  boxed title style={
    outer arc=0pt,
    arc=0pt,
    top=0pt,
    bottom=0pt,
    },
  }
}

\tcbset{warnbox/.style={
  colframe=black,
  colback=yellow!20,
  breakable,
  boxed title style={
    outer arc=0pt,
    arc=0pt,
    top=0pt,
    bottom=0pt,
    },
  }
}
\tcbset{infobox/.style={
  colframe=blue,
  colback=blue!05,
  breakable,
  boxed title style={
    outer arc=0pt,
    arc=0pt,
    top=0pt,
    bottom=0pt,
    },
  }
}

\makeatletter
\def\@LN@depthbox{%
  \ifdim\@tempdima = -1000pt
  \else
    \dp\@tempboxa=\@tempdima
    \nointerlineskip \kern-\@tempdima 
  \fi
  \box\@tempboxa
  } 
\makeatother

\iffancyversion

\newtcbtheorem[crefname={theorem}{theorems},Crefname={Theorem}{Theorems},number within=section]{ttheoremb}{Theorem}{bluewestlinestyle}{r}
\newtcbtheorem[crefname={lemma}{lemmas},Crefname={Lemma}{Lemmas},use counter from=ttheoremb]{tlemmab}{Lemma}{bluewestlinestyle}{r}
\newtcbtheorem[crefname={corollary}{corollaries},Crefname={Corollary}{Corollaries},use counter from=ttheoremb]{tcorollaryb}{Corollary}{bluewestlinestyle}{r}
\newtcbtheorem[crefname={proof}{proofs},Crefname={Proof}{Proofs},number within=section]{tproof}{Proof}{redvioletwestlinestyle}{p}
\newtcbtheorem[crefname={proposition}{propositions},Crefname={Proposition}{Propositions},use counter from=ttheoremb]{tprop}{Proposition}{bluewestlinestyle}{r}
\newtcbtheorem[crefname={assumption}{assumptions},Crefname={Assumption}{Assumptions},use counter from=ttheoremb]{tassumption}{Assumption}{greenwestlinestyle}{a}
\newtcbtheorem[crefname={remark}{remarks},Crefname={Remark}{Remarks},number within=section, ]{tremark}{Remark}{greenwestlinestyle}{rm}
\newtcbtheorem[crefname={example}{examples},Crefname={Example}{Examples},number within=section,]{texample}{Example}{greenwestlinestyle}{ex}
\newtcbtheorem[crefname={definition}{definitions},Crefname={Definition}{Definitions},number within=section]{tdefinition}{Definition}{bluewestlinestyle}{d}

\else

\newtcbtheorem[crefname={theorem}{theorems},Crefname={Theorem}{Theorems},number within=section]{ttheoremb}{Theorem}{defaultthmstyle}{r}
\newtcbtheorem[crefname={lemma}{lemmas},Crefname={Lemma}{Lemmas},use counter from=ttheoremb]{tlemmab}{Lemma}{defaultthmstyle}{r}
\newtcbtheorem[crefname={corollary}{corollaries},Crefname={Corollary}{Corollaries},use counter from=ttheoremb]{tcorollaryb}{Corollary}{defaultthmstyle}{r}
\newtcbtheorem[crefname={proposition}{propositions},Crefname={Proposition}{Propositions},use counter from=ttheoremb]{tprop}{Proposition}{defaultthmstyle}{r}
\newtcbtheorem[crefname={proof}{proofs},Crefname={Proof}{Proofs},number within=section]{tproof}{Proof}{defaultproofstyle}{p}
\newtcbtheorem[crefname={remark}{remarks},Crefname={Remark}{Remarks},number within=section, ]{tremark}{Remark}{defaultthmstyle}{rm}
\newtcbtheorem[crefname={assumption}{assumptions},Crefname={Assumption}{Assumptions},use counter from=ttheoremb, ]{tassumption}{Assumption}{compactdefaultstyle}{a}
\newtcbtheorem[crefname={example}{examples},Crefname={Example}{Examples},number within=section,]{texample}{Example}{compactdefaultstyle}{ex}
\newtcbtheorem[crefname={definition}{definitions},Crefname={Definition}{Definitions},number within=section]{tdefinition}{Definition}{compactdefaultstyle}{d}

\fi

\iffancyversion
	\makeatletter
	\newcommand\linktoproofifdef[1]{%
	  \@ifundefined{r@#1}{}{%
	\null\hfill\hyperref[#1]{\color{customBlue}{($→$ Proof)}}%
	  }%
	}
	\makeatother
\else
	\newcommand\linktoproofifdef[1]{}
\fi

\newif\ifnolinkoption
\NewDocumentEnvironment{tlemma}{omm}{
\IfValueTF{#1}{\IfSubStr{#1}{n}{\nolinkoptiontrue}{\nolinkoptionfalse}
}{ \nolinkoptionfalse }
\begin{tlemmab}[%
	after upper={\ifnolinkoption{}\else{\linktoproofifdef{p:r:#3}}\fi},
	]{#2}{#3}	
}{
\end{tlemmab}
}
\NewDocumentEnvironment{ttheorem}{omm}{
\IfValueTF{#1}{\IfSubStr{#1}{n}{\nolinkoptiontrue}{\nolinkoptionfalse}
}{ \nolinkoptionfalse }
\begin{ttheoremb}[%
	after upper={\ifnolinkoption{}\else{\linktoproofifdef{p:r:#3}}\fi},
	]{#2}{#3}	
}{
\end{ttheoremb}
}
\NewDocumentEnvironment{tcorollary}{omm}{
\IfValueTF{#1}{\IfSubStr{#1}{n}{\nolinkoptiontrue}{\nolinkoptionfalse}
}{ \nolinkoptionfalse }
\begin{tcorollaryb}[%
	after upper={\ifnolinkoption{}\else{\linktoproofifdef{p:r:#3}}\fi},
	]{#2}{#3}	
}{
\end{tcorollaryb}
}

\iffancyversion
	\newenvironment{tproofof*}[1]{
	\begin{tproof}[title={Proof of \Cref{#1}:}]{}{#1}
	}{
	\end{tproof}
	}
\else
	\newenvironment{tproofof*}[1]{
	\begin{tproof}[title={Proof of \Cref{#1}:}]{}{#1}
	}{%
\iffancyversion\else\qed\fi%
\end{tproof}
	}
\fi

\newcommand\proofparagraph[1]{\paragraph{#1}}
\fi

\renewcommand\l\nf
\newcommand\cc[1]{{\smaller #1}}

\crefformat{equation}{eq.~(#2#1#3)}
\Crefformat{equation}{Eq.~(#2#1#3)}

\ifejs
	\newcommand\pcite\citep
	\newcommand\tcite\citet
\else
	\usepackage[style=authoryear-comp, hyperref=true, backend=bibtex,sorting=none,url=false,doi=false,maxcitenames=1,uniquelist=false,sorting=nyt,maxbibnames=8]{biblatex}
	\newcommand\pcite\parencite
	\newcommand\tcite\textcite
	\AtEveryBibitem{
		\clearfield{urldate}
		\clearfield{urlyear}
		\clearfield{urlmonth}
		\clearfield{urlday}
	}
	\AtEveryBibitem{\clearfield{month}}
	\AtEveryCitekey{\clearfield{month}}
	\AtEveryBibitem{\clearfield{day}}	
	\AtEveryCitekey{\clearfield{day}}
	\DeclareSourcemap{
	  \maps[datatype=bibtex]{
		  \map{
			\step[fieldset=urldate, null]
		  }
	   }
	}

	\DeclareRobustCommand{\SkipTocEntry}[5]{}
\fi

\ifejs
	
\else

	\makeatletter
	\def\paragraph{\@startsection{paragraph}{4}%
	  \z@\z@{-\fontdimen2\font}%
	  {\normalfont\bfseries}}
	\makeatother

	\title{A Scalable Nyström-Based Kernel \\Two-Sample Test with Permutations}

	\usepackage{authblk}
	\author[1,2]{Antoine Chatalic}
	\author[1,3]{Marco Letizia}
	\author[1,4]{Nicolas Schreuder}
	\author[1,5,6]{Lorenzo Rosasco}
	\affil[1]{MaLGa Center - DIBRIS, Università di Genova, Genoa, Italy}
	\affil[2]{CNRS, Univ. Grenoble-Alpes, GIPSA-lab, France}
	\affil[3]{INFN - Sezione di Genova, Genoa, Italy}
	\affil[4]{CNRS, Laboratoire d'Informatique Gaspard Monge, Champs-sur-Marne, France}
	\affil[5]{Center for Brains, Minds and Machines, MIT, Cambridge, MA, USA}
	\affil[6]{Istituto Italiano di Tecnologia, Genoa, Italy}
	\date{\today} 
\fi

\begin{document}

\ifejs
	\begin{frontmatter}
	\title{A Scalable Nyström-Based Kernel \\Two-Sample Test with Permutations}
	\runtitle{A Scalable Nyström-Based Kernel Two-Sample Test with Permutations}

	\begin{aug}
	\author[A]{\fnms{Antoine}~\snm{Chatalic}\ead[label=e1]{antoine.chatalic@cnrs.fr}\orcid{0000-0003-2574-2417}},
	\author[B,C]{\fnms{Marco}~\snm{Letizia}\ead[label=e2]{marco.letizia@edu.unige.it}\orcid{0000-0001-9641-4352}}
	\author[D]{\fnms{Nicolas}~\snm{Schreuder}\ead[label=e3]{nicolas.schreuder@cnrs.fr}\orcid{0000-0001-7363-8679}}
	\and
	\author[B,E,F]{\fnms{Lorenzo}~\snm{Rosasco}\ead[label=e4]{lrosasco@mit.edu}\orcid{0000-0003-3098-383X}}

	\address[A]{CNRS, Univ. Grenoble-Alpes, GIPSA-lab, France\printead[presep={,\ }]{e1}}
	\address[B]{MaLGa Center - DIBRIS, Università di Genova, Genoa, Italy\printead[presep={,\ }]{e2,e4}}
	\address[C]{INFN - Sezione di Genova, Genoa, Italy}
	\address[D]{CNRS, Laboratoire d'Informatique Gaspard Monge, Champs-sur-Marne, France\printead[presep={,\ }]{e3}}
	\address[E]{CBMM - Massachusets Institute of Technology, Cambridge, MA, USA}
	\address[F]{Istituto Italiano di Tecnologia, Genoa, Italy}

	\runauthor{A. Chatalic et al.}
	\end{aug}

	\begin{abstract}
\else
	\maketitle
	\paragraph{Abstract}
\fi
Two-sample hypothesis testing–determining whether two sets of data are drawn from the same distribution–is a fundamental problem in statistics and machine learning with broad scientific applications. In the context of nonparametric testing, maximum mean discrepancy (MMD) has gained popularity as a test statistic due to its flexibility and strong theoretical foundations. However, its use in large-scale scenarios is plagued by high computational costs. In this work, we use a Nyström approximation of the MMD to design a computationally efficient and practical testing algorithm while preserving statistical guarantees. Our main result is a finite-sample bound on the power of the proposed test for distributions that are sufficiently separated with respect to the MMD. The derived separation rate matches the known minimax optimal rate in this setting. We support our findings with a series of numerical experiments, emphasizing applicability to realistic scientific data.
\ifejs
	\end{abstract}

	\begin{keyword}[class=MSC]
	\kwd[Primary ]{62G10}
	\end{keyword}

	\begin{keyword}
	\kwd{Nonparametric hypothesis testing}
	\kwd{two-sample testing}
	\kwd{kernel methods}
	\kwd{Nyström approximation}
	\end{keyword}

	\end{frontmatter}
\else
\fi


\section{Introduction}

Let $P,Q$ be probability distributions over a measurable space~$\dsp$. 
We consider the two-sample hypothesis testing problem, where one observes independent random samples
\begin{align}
	X_1, \dots, X_{\nX} \diid P \quad \text{ and } \quad
	Y_1, \dots, Y_{\nY} \diid Q,
	\label{h:iid_data}
\end{align}
and would like to test the hypothesis $\mathcal{H}_0 : P=Q$ against the alternative $\mathcal{H}_1 : P\neq Q$. 

This problem is of paramount importance 
in many scientific areas, particularly in precision sciences such as high-energy physics, where large volumes of multivariate data are commonly analyzed. An example of an application is the search for new physics, where small deviations between observed and predicted distributions may indicate new phenomena, see, \eg, \pcite{Letizia:2022xbe, chakravarti2023model, DAgnolo:2019vbw}. Another important example is the validation of generative models, which are emerging as promising surrogates to replace expensive Monte Carlo simulations while maintaining high fidelity, see \eg \pcite{Grossi:2024axb, Krause:2024avx}.

This paper focuses on kernel-based two-sample tests, which uses the maximum mean discrepancy (MMD; see \Cref{s:mmd}) as a test statistic 
\pcite{gretton2012KernelTwosampleTest}.
These tests are versatile and powerful tools for comparing probability distributions $P$ and $Q$ without requiring strong prior assumptions. Moreover, \new{they enjoy optimal statistical guarantees  with respect to broad classes of alternative distributions~\pcite{schrab2023MMDAggregatedTwoSample, fixed_kim2024DifferentiallyPrivatePermutation}}. However, they are often hindered by their computational cost, which scales quadratically with the total sample size $n \coloneqq \nX + \nY$.

In order to mitigate this computational drawback, several approximations of the MMD have been explored, each offering a trade-off between \new{computational} efficiency and \new{statistical power}. Notable approaches include linear-MMD \pcite{gretton2012KernelTwosampleTest}, incomplete U-statistics \pcite{yamada2018PostSelectionInference,schrab2022EfficientAggregatedKernel}, and block-MMD \pcite{zaremba2013BtestNonparametricLow}. A key limitation of these approaches is that quadratic time complexity is still necessary to achieve statistical optimality \pcite[Proposition 2]{fixed_domingo-enrich2023CompressThenTest}. More details on related works will be provided in Section~\ref{s:related}.

To address this limitation, we propose a new procedure for two-sample testing based on a Nyström approximation of the MMD \pcite{nystroem1930UeberPraktischeAufloesung,williams2001UsingNystromMethod,chatalic2022NystromKernelMean}. From a theoretical perspective, we put forward a setting in which we prove the minimax optimality\new{--in terms of separation rates \pcite{baraud2002nonasymptotic}--}of our procedure while maintaining sub-quadratic computational complexity—a property shared with random feature approximations \pcite{zhao2015FastMMDEnsembleCircular,choi2024computational} and coreset-based methods \pcite{fixed_domingo-enrich2023CompressThenTest}. Moreover, our method is simple to implement, with its approximation quality controlled by a single hyperparameter, making it both computationally efficient and easy to use in practice. Additionally, it applies to a broad class of kernels and does not require the input space to be a Euclidean space. 
The applicability of our approach is illustrated through the example of particle detection on high-energy physics datasets. 

\paragraph{Organization of the paper}
We begin by reviewing the two-sample testing problem and kernel-based two-sample tests in \Cref{sec:background}, followed by related work in \Cref{s:related}. \Cref{sec:test} introduces our MMD-based permutation test.
Theoretical guarantees for this test are provided in \Cref{sec:theory}. 
Finally, numerical studies presented in \Cref{sec:exp} demonstrate the practical effectiveness of our method.

\ifejs\else\Cref{t:notations} contains the main notations used throughout the paper. 
	\begin{table}
	\begin{center}
	\begin{tabular}{@{}ll@{}}
		\toprule
		$\dsp$ & Data space \\
		$\ts$ & Test statistic \\
		$\kr, \fmap$ & (Base) kernel and associated feature map\\
		$\tk, \tfmap$ & Feature map used to build the test statistic and associated kernel\\
		$\thr$ & Threshold of the test \\
		$\nX=|X|, \nY=|Y|$ & Number of samples (from $P$, from $Q$)  \\
		$n=\nX+\nY$ & Total number of samples  \\
		$\nf$ & dimension of the feature map \\
		$\np$ & Number of permutations \\
		\bottomrule
	\end{tabular}
	\end{center}
	\caption{\label{t:notations}Main notations used throughout the paper}
	\end{table}
\fi

\section{Background on testing}\label{sec:background}

In this section, we formally introduce the two-sample testing problem and provide an overview of MMD-based procedures to address it.

\subsection{The two-sample testing problem}
\label{s:testing}

Let $(\dsp, \mathcal{M})$ be a measurable space%
. Let $\cP(\dsp)$ be the space of probability measures on $\dsp$, and consider two probability distributions $P,Q∊\cP(\dsp)$. 
Recall that we are given \tiid random samples 
$X₁,…,X_{\nX}\sim P$ and $Y₁,…,Y_{\nY}\sim Q$ as described in \Cref{h:iid_data}. 
The goal of the two-sample testing problem is to test whether these samples are drawn from the same distribution. Formally, we aim to test the null hypothesis $\mathcal{H}_0 : P=Q$ against the alternative hypothesis $\mathcal{H}_1 : P\neq Q$.

A test is defined as a function of the data $\at:\dsp^{\nX} \times \dsp^{\nY} \to \cb{0,1}$, designed to distinguish between the null and alternative hypotheses. Specifically, the test rejects the null hypothesis if and only if $\at(\X,\Y) = 1$. 
Most tests are built by considering a \textbf{test statistic}~$Ψ$ measuring some dicrepancy between $\X$ and $\Y$, and reject the null if and only if $Ψ(\X,\Y)>t$ for some threshold $t$, which can also depend on $\X,\Y$ and whose choice is part of the definition of the test. 
To evaluate the effectiveness of such a test, we consider the errors it may incur in distinguishing between the null and alternative hypotheses.
\begin{tdefinition}{Type I/II errors}{}
	A type I error occurs when the null hypothesis $P = Q$ is incorrectly rejected. Conversely, a type II error occurs when the null hypothesis $P = Q$ is not rejected, even though $P \neq Q$.
\end{tdefinition}
From these definitions, we can characterize the performance of a test in terms of its level and power.
\begin{tdefinition}{Level, power}{}
	A test $\at$ is said to have level $α$ if its type I error is uniformly bounded by $α$, 
	and power $1-β$ for a class of alternative distributions $\cP₁ \subseteq \cP(\dsp)^2$ if its type II error is uniformly bounded by~$β$: 
	\begin{align}
		\sup_{P=Q ∊ \cP(\dsp)} \P_{\substack{\X \sim P^{\otimes \nX} \\ \Y \sim Q^{\otimes \nY}}}[\at(\X,\Y) = 1] 
		&\leq α, \label{e:def_level}\\
		\sup_{(P,Q)∊\cP_1} \P_{\substack{\X \sim P^{\otimes \nX} \\ \Y \sim Q^{\otimes \nY}}}[\at(\X,\Y) = 0] 
		&\leq β. \label{e:def_power}
	\end{align}
	The probabilities  are evaluated over the data $\X$, $\Y$, and over all other sources of randomness involved in the testing procedure.
\end{tdefinition}

\new{While level and power characterize the performance of a test for a fixed alternative class, they do not describe how the difficulty of the testing problem varies with the separation between the probability distributions $P$ and $Q$.
The minimax framework of 
\pcite{ingster1987minimax,ingster1993asymptotically} addresses this issue by parameterizing the alternative through a separation parameter~$ρ$. In this setting, 
denoting $\cP_1(ρ) \coloneqq \{ (P, Q) \in \cP(\dsp)² : d(P, Q) > \rho \}$ for some specified metric $d$ (\eg, the MMD), 
the \textbf{uniform separation rate}  \pcite{baraud2002nonasymptotic} 
\begin{align*}
	ρ(\at,β,\cP_1) \coloneqq \inf\Set{ρ > 0 | \sup_{(P,Q)∊\cP_1(ρ)} \P_{\substack{\X \sim P^{\otimes \nX} \\ \Y \sim Q^{\otimes \nY}}}[T(\X,\Y) = 0] \leq β}
\end{align*}
provides a concrete characterization of a test's performance by quantifying the smallest separation $\rho$ between distributions $P$ and $Q$ 
under which the test $\at$ achieves a prescribed power $1-\beta$.}
Finally, for a fixed (non-asymptotic) level $α$, the \textbf{minimax rate of testing} is defined as  the smallest 
uniform separation rate achievable by level-$\alpha$ tests:
\begin{align*}
	\underline{ρ}(\nX, \nY, α, β,\cP_1)
	&\de \inf_{T_\alpha}
	ρ(T_\alpha, β, \cP_1),
\end{align*}
where the infimum is taken over all level-$\alpha$ tests $T_\alpha$.

Minimax rates for two-sample testing have been studied under different metrics and various classes of alternative hypotheses, see for instance \tcite{li2024optimality}.
In this paper, we focus on distributions separated with respect to the MMD metric. For this setting, the minimax rate of testing is known to be lower-bounded by  $\log(1 / (\alpha + \beta))^{1/2} \min(\nX, \nY) ^{-1/2}$ for translation-invariant kernels on $ℝ^d$ \pcite[Th. 8]{fixed_kim2024DifferentiallyPrivatePermutation}. In Section~\ref{sec:theory}, we will demonstrate that our proposed test achieves this minimax rate, establishing its optimality with respect to the MMD metric. We now provide a review of kernel-based two-sample tests and related works, before introducing our testing procedure.


\subsection{Kernel-based two-sample tests}
\label{s:mmd}

Let $\delta(\cdot)$ denote the Dirac measure. We define the empirical probability distributions associated with the samples $X$ and $Y$ as $\dX = \frac{1}{\nX} \sum_{i=1}^{\nX} \delta(X_i)$ and $\dY = \frac{1}{\nY} \sum_{i=1}^{\nY} \delta(Y_i)$, respectively. 
A common approach for performing a two-sample test is to assess whether a specific metric between $\dX$ and $\dY$ exceeds a predefined threshold, which typically depends on the sample sizes. In the context of kernel methods, a standard choice of metric is the maximum mean discrepancy between $\dX$ and $\dY$, which we now introduce. 

Let $(\rkhs\new{,\nrkhs{·}})$ be a reproducing kernel Hilbert space (RKHS) with reproducing kernel $\kr:\dsp × \dsp →ℝ$ and canonical feature map $\fmap{x}\de \kr(x,·)$.
We will impose the following assumption on the kernel $\kr$. 
\begin{tassumption}{}{bounded_kernel}
	The kernel is bounded with $\sup_{x∊\dsp}\kr(x,x)=:\supk<∞$ and measurable. 
\end{tassumption}
\new{This assumption is common} in the kernel testing literature (see, e.g., \tcite{fixed_domingo-enrich2023CompressThenTest,choi2024computational}).
The kernel mean embedding \pcite{berlinet2004ReproducingKernelHilbert} of a probability distribution $\new{ν}$ is defined as the Bochner integral \pcite{diestel1977VectorMeasures}
\[ \kme(\new{ν}) \de \int \fmap{x} \dif \new{ν}(x) \new{∊\rkhs}, \]
and is well-defined under \Cref{a:bounded_kernel}. Kernel mean embeddings allow probability distributions over arbitrary spaces to be represented as points in a Hilbert space.
The maximum mean discrepancy (MMD) between two probability distributions $P$ and $Q$ is then defined as the distance between their respective kernel mean embeddings,
\begin{align}
	\MMD(\new{ν}₁,\new{ν}₂) 
	&\de \nrkhs*{μ(\new{ν}₁)-μ(\new{ν}₂)}. \label{e:def_MMD}
\end{align}

A kernel is said to be characteristic \pcite{fukumizu2007kernel} if and only if the mapping $μ$ is injective, \ie 
$P=Q \iff \nrkhs*{μ(P)-μ(Q)}=0$,
in which case the MMD defines a metric on $\cP(\dsp)$. 
Examples of characteristic kernels \new{on $ℝ^d$} include Gaussian, Laplace and Matérn kernels. For general conditions under which kernels are characteristic, see \tcite{sriperumbudur2010RelationUniversalityCharacteristic}, \tcite{simon2018kernel} and related references. 

Kernel two-sample testing is based on the principle that the MMD between two samples drawn from the same distribution should be small. 
The null hypothesis is rejected if the MMD (or a related statistic) exceeds a predefined threshold, indicating that the two samples are likely drawn from different distributions. 
Existing approaches mainly differ in their choice of test statistic and the method used to determine the test threshold.

\section{Related work}\label{s:related}

Hypothesis testing and two-sample testing have been widely studied for a long time, and we refer the reader to \tcite{lehmann_testing_2022} for a general introduction. 

\paragraph{Kernel-based test} 
The introduction of two-sample tests using the MMD and its unbiased estimators as test statistics is due to \tcite{gretton2007KernelMethod,gretton2012KernelTwosampleTest}.
Based on either large deviation bounds or the asymptotic distribution of the unbiased test statistic, the authors derive test threshold values to achieve a target significance level $α$. Following this work, many variants have been proposed. For example, to address the issue that the standard kernel-MMD test statistic is a degenerate U-statistic under the null hypothesis, making its limiting distribution intractable, \tcite{shekhar2022PermutationfreeKernelTwosample} introduced cross-MMD. This quadratic-time MMD test statistic uses sample-splitting and studentization to ensure a limiting standard Gaussian distribution under the null.
Departing from the MMD, other kernel-based test statistics have been explored. 
Since the MMD is an integral probability metric \pcite{muller1997integral}, MMD-based tests can be interpreted as identifying the most discriminative test function from a set of witness functions belonging to a reproducing kernel Hilbert space. Inspired by this interpretation, tests based on optimized witness functions have been proposed by \tcite{kubler2022WitnessTwoSampleTest} and \tcite{kubler2022AutoMLTwoSampleTest}.
Other kernel-based metrics include kernel Fisher discriminant analysis \pcite{harchaoui2008TestingHomogeneityKernel}, and its variants regularized by truncation of the spectrum \pcite{ozier-lafontaine2024KernelbasedTestingSinglecella} or more general spectral filtering \pcite{hagrass2024spectral}. This approach can be viewed as a kernelized version of Hotelling’s $T^2$ test statistic. Additional approaches include the kernel density ratio \pcite{kanamori2011DivergenceEstimationTwoSample}, kernel Stein discrepancies for goodness-of-fit tests \pcite{huggins2018RandomFeatureStein,kalinke2024nystr}, 
\new{and test statistics based on covariance operators for independence testing \pcite{zhang2018LargescaleKernelMethods}.}

\paragraph{Efficient kernel-based tests}
The main disadvantage of kernel-based tests is that computing the MMD scales quadratically with the number of samples $n$.
In their seminal paper, \tcite[Section 6]{gretton2012KernelTwosampleTest} already introduced the linear-MMD, a statistic computable in $O((n+m)d)$ time, leveraging a partial evaluation of the terms appearing in the U-statistic estimator of the squared MMD. 
Variants of these incomplete U-statistics have subsequently been proposed by \tcite{yamada2018PostSelectionInference,schrab2022EfficientAggregatedKernel}. 
Considering another partial evaluation of the MMD, \tcite{zaremba2013BtestNonparametricLow} introduced the block-MMD, a test statistic derived by (i) splitting the observations into disjoint blocks, (ii) computing the MMD for each block, and (iii) averaging the resulting statistics across all blocks. This approach has been further analyzed and refined by \tcite{ramdas2015adaptivity} and \tcite{reddi2015high}.
\tcite{chwialkowski2015FastTwosampleTesting} introduced a linear-time test statistic based on the average squared distance between empirical kernel embeddings evaluated at~$J$ randomly drawn points. \tcite{jitkrittum2016InterpretableDistributionFeatures} proposed a variant of this statistic in which the~$J$ points are selected to maximize a lower bound on the power.
A major limitation of these approaches is that either a quadratic time complexity is necessary to achieve an optimal power \pcite[Proposition 2]{fixed_domingo-enrich2023CompressThenTest} or the computation/power trade-off is yet to be characterized.
Coreset-based approximation strategies have also been investigated, and proven to reach minimax separation rates at a subquadratic computational cost \pcite{fixed_domingo-enrich2023CompressThenTest}. 
Approximations of the MMD based on random Fourier features have been explored \pcite{zhao2015FastMMDEnsembleCircular,zhao2021ComparingDistributionsMeasuring,choi2024computational}. More recently, a test statistic inspired by kernel Fisher discriminant analysis but based on random Fourier features has been introduced, with minimax-optimality guarantees for alternative classes defined via fractional powers of an integral operator~\pcite{mukherjee2025MinimaxOptimalKernel}. 
Finally, \tcite{chatalic2022NystromKernelMean} proposed a Nyström approximation of the MMD,
which is the base of our method and study.


\paragraph{Permutation tests}
There are several popular approaches to determining the test threshold. One common method is to use the quantile of the asymptotic distribution of the test statistic under the null hypothesis. However, this approach provides only asymptotic guarantees and may not perform well in finite samples. Another method relies on concentration inequalities, which, while theoretically sound, can be overly conservative, leading to thresholds that are too loose. Alternatively, permutation and bootstrap methods offer data-driven approaches that approximate the null distribution more accurately in practice, often resulting in improved empirical performance. The idea of comparing test statistics to their permuted replications dates back to the work of \tcite{hoeffding1952large}. More recently, the combination of the MMD test statistic with permutation and bootstrap approaches has been explored in works such as \tcite{fromont2012KernelsBasedTests,kim2022MinimaxOptimalityPermutationa,schrab2023MMDAggregatedTwoSample}.

\paragraph{Parameter selection} 
The kernel function and its hyperparameters play an important role in the practical usability of kernel-based tests. 
\tcite{balasubramanian2021OptimalityKernelembeddingBased} and \tcite{li2024optimality} established minimax optimality of carefully tuned MMD-based tests in specific settings.
Multiple approaches have been investigated, such as aggregating multiple tests~\pcite{fromont2012KernelsBasedTests,schrab2023MMDAggregatedTwoSample,schrab2022EfficientAggregatedKernel,biggs2023MMDFUSELearningCombining} or using a Bayesian formalism \pcite{zhang2022BayesianKernelTwoSample}.
In this work, we focus on the computational efficiency of the test, yet our approach could easily be combined with such ideas when adaptation is needed. 
\new{We will also see in \Cref{sec:theory} that the choice of the kernel has direct implications on the approximability of the MMD.}

\section{An approximate MMD permutation test}
\label{sec:test}

As noted in the previous section, computing the MMD between two empirical distributions has  \(O(n^2)\) \new{runtime} complexity since it requires computing all pairwise kernel evaluations, making it impractical for large datasets. 
To address this limitation, we introduce an efficient randomized approximation of the MMD that we will use as a test statistic.

\subsection{Projection-based approximation of the MMD}

We approximate the MMD \eqref{e:def_MMD} between two empirical distributions $\hat{P}$ and $\hat{Q}$ using the  Nyström method \pcite{nystroem1930UeberPraktischeAufloesung,williams2001UsingNystromMethod}, that is by projecting their kernel mean embeddings $\mu(\hat{P})$ and $\mu(\hat{Q})$ onto a data-dependent finite-dimensional subspace.
\new{Here and throughout the paper, for any $h_1,…,h_ℓ∊\rkhs$ we identify $[h_1,…,h_l]$ with the (finite-rank) column operator $x∊ℝ^ℓ↦\sum_{i=1}^ℓ x_i h_i∊\rkhs$.}
In particular, given $\nf \in \mathbb{N}^*$ landmark points $\ldm{1},…,\ldm{\nf}∊\dsp$,
we define the operator $\ftldms=[{\fmap{\ldm{1}},…, \fmap{\ldm{\nf}}}]:ℝ^{\nf}→\rkhs$ and introduce the orthogonal projector $\Pm:\rkhs→\rkhs$ onto $\spa(\ftldms)$. We then approximate  $\MMD(\hat{P}, \hat{Q})$ by 
\begin{align}
	\ts 
	&\de \nrkhs*{ \Pm μ(\dX) - \Pm μ(\dY) }.
	\label{e:def_ts} 
\end{align}

We now derive a more practical form of this expression for efficient implementation.
\new{We denote $\ftldms^*:\rkhs→ℝ^{\nf}$ the adjoint of $\ftldms$, which can be written $\ftldms^*:h \mapsto [h(\ldm{1}),…, h(\ldm{\nf})]ᵀ$ using the reproducing property of the RKHS.}
The projector $\Pl$ can then be expressed as $\Pl=\ftldms(\ftldms^*\ftldms)^†\ftldms^*$. Note that it satisfies the standard projection property $\Pl=\Pl^2=\Pl^*$.
We also stress that $\ftldms^*\ftldms=\Kl$ corresponds to the kernel matrix 
of the landmarks, \ie, 
$(\Kl)_{ij}=\kr(\ldm{i},\ldm{j})$ for $i,j=1, \dots, \nf$. 
Hence, for any $v∊\rkhs$, it holds $\nrkhs{\Pm v}^2=\ip{v,\Pl^2 v}=\ip{v,\Pl v}=\n{(\Kl^†)^{1/2}\ftldms^* v}$. 
In particular, this implies that the approximation of the MMD in \Cref{e:def_ts} can be computed efficiently as
\begin{align}
	\ts
	&= \n*{ \tfrac{1}{\nX} \sum_{i=1}^{\nX} \tfmap(x_i) - \tfrac{1}{\nY} \sum_{j=1}^{\nY} \tfmap(y_j) }
	\label{e:def_ets} 
	\\
	\quad\text{where}\quad
	&\tfmap(x) 
	\de (\Kl^†)^{1/2}\vvec{κ(\ldm{1}, x) \\ …\\κ(\ldm{\nf}, x) }\new{∊\bR^{\nf}},
	\label{e:def_fmap} 
\end{align}
and $\Kl^†$ denotes the Moore-Penrose pseudo-inverse of $\Kl$. 
Note that $\ts$ can be computed in a single pass over the data once the landmarks are fixed. \new{More precisely, given the landmarks and $\Kl^†$, evaluating $\ts$ requires $n \ell$ kernel evaluations and $O(n \ell^2)$ operations.}

We now detail the procedure used to select these landmarks.

\paragraph{Choice of the landmarks} 
We construct our approximation by sub-sampling $\nf$ points from the pooled dataset $\W=(\XY_i)_{1≤i≤n}$, where $\XY_{i} \coloneqq X_i$ for $1≤i≤\nX$ and $\XY_{\nX+j} \coloneqq Y_{j}$ for $1≤j≤\nY$. 
%
This strategy differs from \tcite{chatalic2022NystromKernelMean}, 
where a similar Nyström approximation is designed by sampling points from both $\X$ and $\Y$ separately. 
Although the two approaches yield similar empirical performance, sampling once from the pooled dataset \new{yields landmarks whose distribution is invariant to permutations of the observations under the null hypothesis, allowing us} to derive a bound on the level of the test with the permutation approach introduced in Section~\ref{subsec:perm}. 


Theoretical guarantees for approximating the MMD using Nyström approximation with uniform sampling of the landmarks from both datasets have been established in \tcite{chatalic2022NystromKernelMean}.  
However, achieving the optimal $O(n^{-1/2})$  rate under uniform sampling typically requires a large number of landmarks \new{or strong smoothness assumptions on the kernel/data}.
In this paper, we instead  adopt leverage scores sampling, which is known to be optimal for approximating  a single kernel mean embedding \new{under mild smoothness assumptions} \pcite{chatalic2025EfficientNumericalIntegration}. 

Leverage scores quantify the relative importance of each point in a dataset and are closely related to the inverse of the Christoffel function in approximation theory~\pcite{fanuel2022NystromLandmarkSampling,pauwels2018RelatingLeverageScores}.
In this work, we focus on kernel ridge leverage scores (KRLS) \pcite{alaoui2015FastRandomizedKernel}. \new{For the pooled dataset $\W$} of size $n$ with associated kernel matrix $K$ \new{defined by $K_{ij}\de \kr(\XY_i,\XY_j)$ for $1≤i,j≤n$}, KRLS are defined as
\begin{equation}
    \tls{i} \coloneqq \prt*{ K(K + λn I)^{-1} }_{ii}, \quad i=1,\dots, n,
	\label{e:def_ls}
\end{equation}
where $\lambda>0$ is a regularization parameter. 
The cost of exactly computing leverage scores quickly becomes prohibitive as the sample size grows due to the matrix inversion. 
Since computing these scores exactly is typically expensive, we consider using multiplicative approximations of these scores.
\begin{tdefinition}{AKRLS}{leverage_scores}
Let $δ \in (0,1]$, $λ_0 > 0$ and $z \in [1 , \infty)$. 
The scores $(\als(i))_{i\in[n]}∊ℝ^n$ are said to be $(z,λ_0,δ)$-approximate kernel ridge leverage scores (AKRLS) of $\X$ if with probability at least $1-δ$, for all $λ≥λ₀,i∊\irange{n}$ it holds $\frac{1}{z}~\tls{i} ≤ \als{i}≤ z~\tls{i}$.
\end{tdefinition}
Efficient algorithms have been proposed in the literature to efficiently sample from such approximate kernel ridge leverage scores, see for instance \tcite{musco2017RecursiveSamplingNystrom,rudi2018FastLeverageScore,chen2021FastStatisticalLeverage}. \new{We discuss the time complexity of this sampling step at the end of the section.}

\new{Eventually, we stress that even though our estimator relies on a projection on a subspace spanned by a random subset of the points, it still implicitly depends on all $n$ points of the dataset through the empirical kernel mean embeddings appearing in \eqref{e:def_ets}. This differs notably from coresets approaches, which also rely on a (possibly random) subset of the data, but for which the estimator then only depends on this subset and not on the whole dataset.}

\subsection{Using permutations to determine the threshold}\label{subsec:perm}

To construct a test based on the statistic given in \Cref{e:def_ts}, we must specify how to choose the test threshold. If the distribution of the test statistic under the null was known, we would select the quantile corresponding to the desired test level. However, in general, the exact distribution of the test statistic is unknown and intractable. Furthermore, since we seek finite-sample guarantees, we avoid relying on limiting distributions.
Instead, we adopt a permutation-based approach \pcite[Chapter 17]{lehmann_testing_2022}. The core idea of this approach is to exploit the fact that, under the null hypothesis (i.e., when $P=Q$), all observed data points are exchangeable. As a result, permuting the samples leaves the distribution of the test statistic unchanged and enables to approximate its distribution.

%
In what follows we consider a random variable $σ$ that is  uniformly distributed over the set $\new{S_n}$ of all permutations of $\cb{1,…,n}$ and that is independent of all other sources of randomness. 
By sampling $\np$ \tiid permutations $(σ_p)_{p=1}^\np$ and taking $σ_0=\textup{Id}$, we compute $(\np+1)$ permuted test statistics
\[ 
\ipts{p} \de \ts[(\XY_{σ_p(i)})_{i=1}^{n_X},(\XY_{σ_p(j)})_{j=n_X+1}^{n}],\quad 0≤p≤\np,
\]
\new{where we recall that $\XY$ denotes the pooled dataset.}
We use the notation $\oipts{i}$ to refer to the $i$-th value of the \emph{ordered} list of these statistics, \ie $\oipts{0}≤…≤\oipts{p}$.  
The threshold is then set as the empirical quantile $\oipts{b_α}$ of the permuted test statistics, where $b_α\de \lceil (1-α)(\np+1) -1\rceil$.
Our testing procedure is detailed in \Cref{al:main_test}. We next discuss its computational and memory efficiency, before presenting statistical guarantees in the following section.


\newcommand\mycommfont[1]{{\smaller #1}}
\begin{algorithm2e}
	\KwIn{Feature map $\varphi$ as in \Cref{e:def_fmap},
			data
			$\XY=(\XY_i)_{1≤i≤n}∊\dsp^n$, 
			level~$α \in (0, 1)$,
			number of permutations~$\np$}
	\KwOut{Result of the test (boolean)}
	$w^{(0)} ← \brk*{\tfrac{1}{\nX}, …, \tfrac{1}{\nX}, -\tfrac{1}{\nY}, …, -\tfrac{1}{\nY} }∊ℝ^n$ \; 
	\lForEach{$p=1,…,\np$}{
		$w^{(p)} ← \textup{shuffle}(w)$ 
	}
	\lForEach{$p=0,…,\np$}{
		$v^{(p)} ← 0∊ℝ^{\nf}$ 
	}
	\ForEach(\tcp*[f]{$O(n\np \nf)$ time, $O(\np\nf)$ space}){$i=1,…,n$}{%
		\lForEach{$p=0,…,\np$}{%
			$v^{(p)} ← v^{(p)} + w^{(p)}_i \tfmap(\XY_i)$ %
		}%
	}
	\lForEach{$p=0,…,\np$}{
		$\ipts{p} ←\n*{v^{(p)}}$ 
	}
	$b_α ← \lceil (1-α)(\np+1) -1\rceil$\;
	\uIf{$\ipts{0} > \oipts{b_α}$}{
		return $1$
			\tcp*{reject $H₀$}
	}\uElseIf{$\ipts{0} = \oipts{b_α}$}{
		$\tsl^> ← \#\{0 ≤ b ≤ \np : \ipts{b} > \oipts{b_α}\}$\;
		$\tsl^= ← \#\{0 \leq b ≤ \np : \ipts{b} = \oipts{b_α}\})$\;
		return $1$ with probability $\frac{\alpha(\np+1) - \tsl^>}{\tsl^=}$\;
	}\Else{ 
		return $0$ 
		\tcp*{fail to reject $H₀$}
	}
	\caption{Permutation test based on a Nyström approximation of the MMD%
		\label{al:main_test}}
\end{algorithm2e}

\paragraph{Computational and memory efficiency} \new{The approximate feature map used for the computation of our test stastistic allow us to keep} the space and time complexity of our algorithm limited. 
Once the landmarks have been chosen, all test statistics are computed in a single pass over the data \new{for a total $O(n \nf (c_κ + \np))$ runtime, where $c_κ$ denotes the cost of a kernel evaluation. One must add to this cost the computations involved in the definition of the feature map $φ$, 
which includes $O(\nf³)$ for the computation of the pseudo-inverse in $\eqref{e:def_fmap}$, 
and the cost of sampling the landmarks, which depends on the algorithm used%
\footnote{\new{In the experimental section, we rely for simplicity on \tcite[Algorithm 3]{musco2017RecursiveSamplingNystrom}, whose time complexity scales in $O(n \nf²)$, however other algorithms such as \tcite{rudi2018FastLeverageScore} could be used.}}.}

Moreover, the~$\np$ permutations can be efficiently stored using $O(\np n)$ bits and only \new{$O((\np+1) \nf)$} space is required to store the mean embeddings corresponding to all permutations and the non-permuted test statistic. 
In particular, it is never needed to store the features of the whole dataset in memory. 
This enables us to easily permute all samples without resorting to batch compression strategies that are typically required in coresets-based methods~\pcite{fixed_domingo-enrich2023CompressThenTest}.

\section{Statistical guarantees}\label{sec:theory}

\newcommand\cXY{r}
In this section, we provide theoretical guarantees on the level and power of the test described in the previous section. The main results are stated in Section~\ref{subsec:statement}, while the main ingredients for the proofs are outlined in Section~\ref{subsec:proof_elements}. Full proofs are deferred to the appendix.
For simplicity, and without loss of generality, we assume throughout this section that $\cXY=\nX/\nY≤1$. 

\subsection{Level and power guarantees}
\label{subsec:statement}

\paragraph{Level guarantee.}
The next result follows closely \tcite[Theorem 2 and Proposition 3]{hemerik2018ExactTestingRandom}. In our setting, however, additional care is required since the test statistic depends on the landmarks~$\ldms$, which are themselves sampled from the pooled dataset $\W$. The proof is provided in \Cref{s:proof_level}.
\begin{tlemma}{}{level_hemerik_th2}
	Let $α∊[0,1)$ be the desired test level and define
	$b_α\de \lceil (1-α)(\np+1) -1\rceil$.  
	Then \Cref{al:main_test}, when run with input level $α$, has exact rejection probability $α$ under the null hypothesis.
\end{tlemma}
By \emph{exact}, we mean that the inequality in \Cref{e:def_level} is an equality. This exactness is achieved through \new{a simple randomization trick which breaks potential ties (when $\ipts{0} = \oipts{b_α}$) in the algorithm, see \cite[Proposition 3]{hemerik2018ExactTestingRandom}}.
We stress that although the result is stated for \Cref{al:main_test}, it holds more generally when the landmarks are sampled independently and identically from the pooled dataset~$\W$ 
according to probabilities that are equivariant under permutations of the dataset~$\W$.
This assumption is satisfied by both uniform sampling and sampling based on AKRLS computed on the pooled dataset~$\W$. In contrast, sampling landmarks separately from $\X$ and $\Y$ would not satisfy this assumption
Finally, this lemma holds under the assumption that the data are exchangeable under the null hypothesis. This holds in particular when the data are i.i.d., as is the case here, though the \tiid assumption is not explicitly used at this stage.

\paragraph{Power guarantee.} 
We now state our main theoretical result: a bound on the power of the test, whose proof relies on \Cref{r:bound_threshold,r:bound_TS_quantile_rough,r:bound_MMD_nys_new}.

\begin{ttheorem}{Main result: power of the test}{power_MMD}
	Let $\beta\in (0, 1)$ and suppose $1/(\np+1)  \leq α \leq 1/(2e)$. 
	Let $c_α=\lfloor α(\np+1) \rfloor$. 
	Assume that the conditions of \Cref{r:bound_MMD_nys_new} hold with $δ\de β/2$ \new{(that is, the landmarks are chosen using AKRLS sampling, and the number $\nf$ of landmarks is larger than a threshold which depends on regularity assumptions that are made clear below)}. 
%
	Then, there exists a universal constant $c$ such that 
	the test of \Cref{al:main_test} has power at least $1-\beta$, provided that 
	\begin{align*}
		\MMD(P,Q)²
			&≥ \tfrac{c\supk }{\nX} \prt*{ \log\prt*{\frac{2e}{α} \prt*{\frac{2}{β}}^{1/c_α}} 
			+ \log\prt*{\frac{16}{β}} }.
	\end{align*}
\end{ttheorem}
A \new{slightly weaker separation condition}, with constants that are made comparatively more explicit is provided in the proof of Theorem~\ref{r:power_MMD}. \new{This finer bound also explicitely depends on the ratio $\cXY=\nX/\nY$, while we used here $\cXY≤1$ for simplification.}
Note that the dependence of the separation rate (in terms of the MMD) on the smallest sample size $\nX$  is $\nX^{-1/2}$. It matches the known minimax optimal rate of testing in this setting, as established in \Cref{s:testing} \pcite[Th. 8]{fixed_kim2024DifferentiallyPrivatePermutation}.

\subsection{Main ingredients for the power guarantee}
\label{subsec:proof_elements}

The starting point of our proof is the following lemma, which is \new{a generalization of} \tcite[Lemma \new{2}]{schrab2023MMDAggregatedTwoSample}. It shows that a lower bound on the power can be obtained from a high-probability control of the MMD estimation error.

\begin{tlemma}{}{bound_threshold}
	Let $0 < \beta \leq 1$, and let $X$ and $Y$ be datasets of sizes $\nX$ and $\nY$, sampled from distributions $P$ and $Q$, respectively.
	Assume that there exists a function $\eMMDv$ such that an estimator $\ts$ of the MMD satisfies 
	\begin{align}
		\P[|\MMD(P,Q)-\ts|≥\eMMD[β/2]] &≤ β/2. 
		\label{e:hyp_mmd_quality}
	\end{align}
	If the distributions $P$ and $Q$ are such that
	\begin{align*}
		\P[\MMD(P,Q)≥\eMMD[\nicefrac{\beta}{2}] +\thr] > 1-\beta/2\enspace,
	\end{align*}
	then $\P[\ts≤\thr]≤β$. 
\end{tlemma}

In \Cref{s:mmd_approximation}, we will show that \Cref{e:hyp_mmd_quality} is satisfied for the proposed Nyström-based estimator. 
Then, following 
 \tcite[Lemma 6]{fixed_domingo-enrich2023CompressThenTest} 
we will show in \Cref{s:bound_quantile} how the (random) empirical quantile threshold $\thr$ can be bounded by the (deterministic) quantile of the permuted test statistic, conditionally on $\X,\Y$. 

\subsubsection{Quality of the MMD approximation}
\label{s:mmd_approximation}

The test statistic defined in \Cref{e:def_ts} is an estimator of the MMD between \(P\) and \(Q\). In this section, we provide a high-probability bound for this approximation. This result is of independent interest, as it improves upon \tcite{chatalic2022NystromKernelMean} by using approximate leverage scores sampling instead of uniform sampling.

For $\square \in \{P, Q\}$, let $C_\square \de\E_{x\sim \square}\, ϕ(x)\kron ϕ(x):\rkhs→\rkhs$ denote the (uncentered) covariance operators associated with $P$ and $Q$, \new{where $a\kron b:c↦\ip{b,c}_{\rkhs}\, a$}. Define the pooled covariance operator as $\covM\de \frac{\nX}{n} \covP + \frac{\nY}{n} \covQ $. 
Under \Cref{a:bounded_kernel}, both $C_P$ and $C_Q$ are self-adjoint trace-class operators, and the same holds for $\covM$.
Given a trace-class operator $C$, we denote by $λ_i(C)$ its $i$-th eigenvalue, for \new{$i∊\bN_{>0}$}. We make the following assumption on the spectral decay of the pooled covariance operator.
\begin{tassumption}{Polynomial spectral decay}{poly_decay_mean}
	There exist $γ∊(0,1]$ and $a_γ>0$ such that 
	$λ_i(\covM) ≤ a_γ i^{-1/γ}$.\\
\end{tassumption}
Note that, by Weyl's inequality\footnote{Weyl's perturbation inequality can be easily proved by applying the Courant–Fischer–Weyl min-max principle, which holds for compact self-adjoint operators on a Hilbert space.}, it holds for any $i$ that
\begin{align*} 
	λ_{2i}(\covM)≤λ_{2i-1}(\covM)≤\tfrac{\nX}{n}λ_i( \covP) + \tfrac{\nY}{n}λ_i( \covQ).
\end{align*}
Hence, if both $\covP$ and $\covQ$ exhibit a polynomial spectral decay of the form 
\[\max(λ_i(\covP),λ_i(\covQ)) ≤ a_γ i^{-1/γ}\] for some $γ∊(0,1]$ and $a_γ>0$, then it follows that 
\begin{align*}
	λ_{2i}(\covM) ≤ λ_{2i-1}(\covM) ≤ a_γ i^{-1/γ} = a_γ' (2i)^{-1/γ},
\end{align*}
with $a_γ'\de a_γ 2^{1/γ}$. That is, $\covM$  inherits the same spectral decay rate as $\covP$ and $\covQ$, up to a multiplicative constant.
Assuming that a covariance operator has such a polynomial decay is standard in the literature.
\new{This is for instance the case when considering the Matérn kernel: in this case, the RKHS is norm-equivalent to a Sobolev space, see for instance \tcite{wendland2004ScatteredDataApproximation}, and the eigenvalue decay has been studied for a bounded domain by \tcite{widom1964AsymptoticBehaviorEigenvalues}. } 

A key quantity in the analysis is the so-called effective dimension. This quantity depends both on the choice of the kernel as well as on the probability distributions. We define it for the pooled dataset and any $λ>0$ as 
\begin{align*}
	\mdeff\de \Tr(\covM(\covM+λI)^{-1}).
\end{align*}
Notably, the effective dimension depends on the distributions $P,Q$ only through the covariance operator $\covM$, and can be interpreted as a smooth estimate of the number of eigenvalues of $\covM$ that exceed a given threshold $λ$.
Under \Cref{a:bounded_kernel,a:poly_decay_mean}, there exists a constant $c_γ =c_\gamma (a_γ, \supk)$ such that it holds $\mdeff ≤ c_γ \lambda^{-γ}$ for any $λ>0$ (see \tcite[Section F.1]{chatalic2025EfficientNumericalIntegration}).

We are now ready to state the following result regarding the error induced by the Nyström approximation of the $\MMD$.

\begin{tlemma}{Nyström MMD Approximation}{bound_MMD_nys_new}
	Let $\delta \in (0, 1)$, $z \geq 1$, and $\lambda_0 > 0$.
	Consider $\nf \geq 1$ landmarks $\ldm{1}, \ldots, \ldm{\nf}$ sampled with replacement from the pooled dataset $\XY$ according to the distribution induced by $(z, \lambda_0, δ/6)$-approximate kernel ridge leverage scores (AKRLS). 
	Granted \Cref{a:bounded_kernel,a:poly_decay_mean}, and provided that
	\begin{align*}
		\nf ≥ 4 \left( 1 + \frac{2.4 z² c_γ}{\prt*{16\supk \log(4/δ)}^{γ}} n^γ \right) \log\left(\frac{12n}{δ}\right)  \quad \text{ and } \quad 
		λ_0 ≤ \frac{16\supk\log(4/δ)}{n} ≤ \noprkhs{\covM},
	\end{align*}
	it holds with probability at least $1 - \delta$ that
	\begin{align*}
		|\MMD(P, Q) - \ts| 
		&\leq 10 \supfmap \sqrt{\log(8/δ)} \left( \frac{1}{\sqrt{\nX}} + \frac{1}{\sqrt{\nY}} \right)
			=: \eMMD.
	\end{align*}
\end{tlemma}

This result can be plugged into \Cref{e:hyp_mmd_quality} from \Cref{r:bound_threshold} to obtain a bound on the power. To fully characterize the separation rate, it remains to derive a deterministic bound on the (random) quantile of the permuted test statistic, which is the focus of the next section.

\subsubsection{Bound on the empirical quantile threshold}
\label{s:bound_quantile}

We now derive an upper-bound on the quantile of the permuted test statistic $\pts$ conditionally on $\X,\Y$. We recall that the permutation $σ$ is uniformly distributed over the set $\new{S_n}$ of all permutations of $\{1, \dots, n\}$. 
It will serve as an upper-bound on the random empirical quantile threshold $\thr$ in our main result, which is based on the $\cP$ randomly sampled permutations.
%
For this, we show (cf. \Cref{s:decomposition_U_R}) that the randomly permuted squared test statistic can be 
%
decomposed as the sum of a (weighted) U-statistic and a remainder term:
\begin{align*}
	\pts² 
	&= \tfrac{(\nX-1)(\nY-1)}{\nX\nY} \pstsU + \pstsR.
\end{align*}
%
%
%

%
We bound the quantile of $\pts$ conditionally on $(\X,\Y)$ leveraging a result from \tcite{kim2022MinimaxOptimalityPermutationa} to control the quantile of the $U$-statistic.

\begin{tlemma}{Quantile bound for $\pts |\X,\Y$}{bound_TS_quantile_rough}
Let $0<α≤e^{-1}$. 
Under \Cref{a:bounded_kernel}, there is a universal constant $C'$ such that the test statistic associated to the Nyström kernel approximation satisfies, with probability at least $1-α$, conditioned on the datasets $X$ and $Y$,
\begin{align*}
	\pts^2 
	&≤ C' \tfrac{\supk}{\nX} \log(1/α) + (\tfrac{1}{\nX} +\tfrac{1}{\nY}) 4\supk 
\end{align*}
\end{tlemma}

We use this bound to establish a deterministic upper bound on the (random) empirical quantile threshold $\thr$ using a result from \tcite{fixed_domingo-enrich2023CompressThenTest}.

\begin{tlemma}{Bound on the empirical quantile}{bound_quantile_rough}
	Let $β>0$, and $1/(\np+1) \leq α \leq 1/(2e)$. 	
	Let $c_α=\lfloor α(\np+1) \rfloor$. 
	Then, with probability $1-\beta/2$ conditionally on $X$ and $Y$,
	\begin{align*}
		\thr 
		&≤ \supfmap \left(\tfrac{\sqrt{C'}}{\sqrt{\nX}} \sqrt{\log\prt*{\frac{2e}{α (\beta/2)^{1/c_α}}}} + 2\prt*{\frac{1}{\sqrt{\nX}} +\frac{1}{\sqrt{\nY}}}\right).
	\end{align*}
\end{tlemma}

Using \Cref{r:bound_quantile_rough}, we can now complete the proof of our main result. The full details are deferred to \Cref{s:proof_main_result}.

\section{Numerical studies}\label{sec:exp}

In this section, we explore the empirical power, computational trade-offs, and type-I error rates of our method, comparing against the random Fourier features (RFF) approach of~\cite{choi2024computational} and the compress-then-test (CTT) method of~\cite{fixed_domingo-enrich2023CompressThenTest}.

Throughout our experiments, we employ a Gaussian kernel 
$κ(x,y)=\exp(-‖x-y‖^2/(2h^2)),$
where the bandwidth $h$ is set to the median inter-point Euclidean distance, estimated from a subset of 2000 randomly selected instances.

\new{Regarding landmark selection, we consider both uniform sampling and AKRLS using the recursive ridge leverage score sampling method of~\cite{musco2017RecursiveSamplingNystrom}. For AKRLS we rely on the official implementation\footnote{\url{https://github.com/axelv/recursive-nystrom/tree/master}}, which only requires specifying the number of landmarks and the kernel parameters.}

\new{For the CTT test, we follow the indications of the authors' implementation\footnote{\url{https://github.com/microsoft/goodpoints}}, which requires the resulting bin size to divide each sample size exactly. We then choose the number of bins per sample $s$ as the divisor of $n_X$ closest to $\sqrt{n_X}$ (with $n_X = n_Y$). This ensures that the bin sizes remain integer-valued while yielding an $s$ that scales as $\sqrt{n_X}$.}

We set the test level at $\alpha = 0.05$ and estimate the null distribution of the test statistic using $\mathcal{P} = 199$ permutations. Power estimates are averaged over 400 repetitions, while type-I error rates are averaged over 1000 repetitions. \new{Error bars represent 95\% Wilson score confidence intervals for the underlying Bernoulli success probability \pcite{newcombe1998two}.} Our implementation is open-source and available in Python.
\footnote{\url{https://github.com/mletizia/nystrom-mmd}}

\subsection{Datasets}\label{s:datasets}

\label{p:cg}
\paragraph{Correlated Gaussians} We examine a synthetic dataset comprising 3-dimensional correlated Gaussian distributions. The first sample is drawn from $P = \mathcal{N}_3(\mathbf{0}, \mathbf{\Sigma}(\rho_1))$, where the covariance matrix is defined as \new{$\mathbf{\Sigma}(\rho_1) = \mathbf{I}_3 + \rho_1(\mathbf{1}\mathbf{1}^\top - \mathbf{I}_3)$}, with $\rho_1 = 0.5$. The second sample follows an identical distribution but with varying correlation coefficient $0.51 \leq \rho_2 \leq 0.66$. Sample sizes are fixed at $n_x = n_y = 2500$.


\label{p:higgs}
\paragraph{Higgs and Susy} 
These datasets consist of Monte Carlo simulations of high-energy physics collider data, originally presented in~\cite{Baldi:2014kfa}, characterized by two classes: background data ($P_b$) and signal data ($P_s$). The former represents the prediction of the Standard Model of Particle Physics, the current best description of physical processes at the fundamental level (see \cite{Schwartz:2014sze,ParticleDataGroup:2024cfk}). The signal data represents processes producing new theoretical Higgs bosons and supersymmetric particles for the Higgs and Susy datasets, respectively. 

The Higgs dataset contains 21 features, where the final 7 features are functions of the first 14 and constitute more discriminative \emph{high-level} features. We refer to the initial 14 as \emph{low-level} features. The dataset encompasses 11M examples in total. We focus exclusively on low-level features ($d=14$). \new{The Susy dataset is composed of 8 low-level and 10 high-level features (excluded from our analysis). This dataset encompasses 5M examples in total.}

\new{To replicate typical HEP scenarios and increase test difficulty, we compare data from distribution $P=P_b$ against a mixture of background and signal processes $Q=(1-\alpha_{\rm mix}) P_b + \alpha_{\rm mix} P_s$, with $\alpha_{\rm mix}=0.2$ for the Higgs data and $\alpha_{\rm mix}=0.05$ for the Susy data.}

\subsection{Results}

\begin{figure}
	\centering
	\subfigure
	{%
		\includegraphics[width=0.48\linewidth]{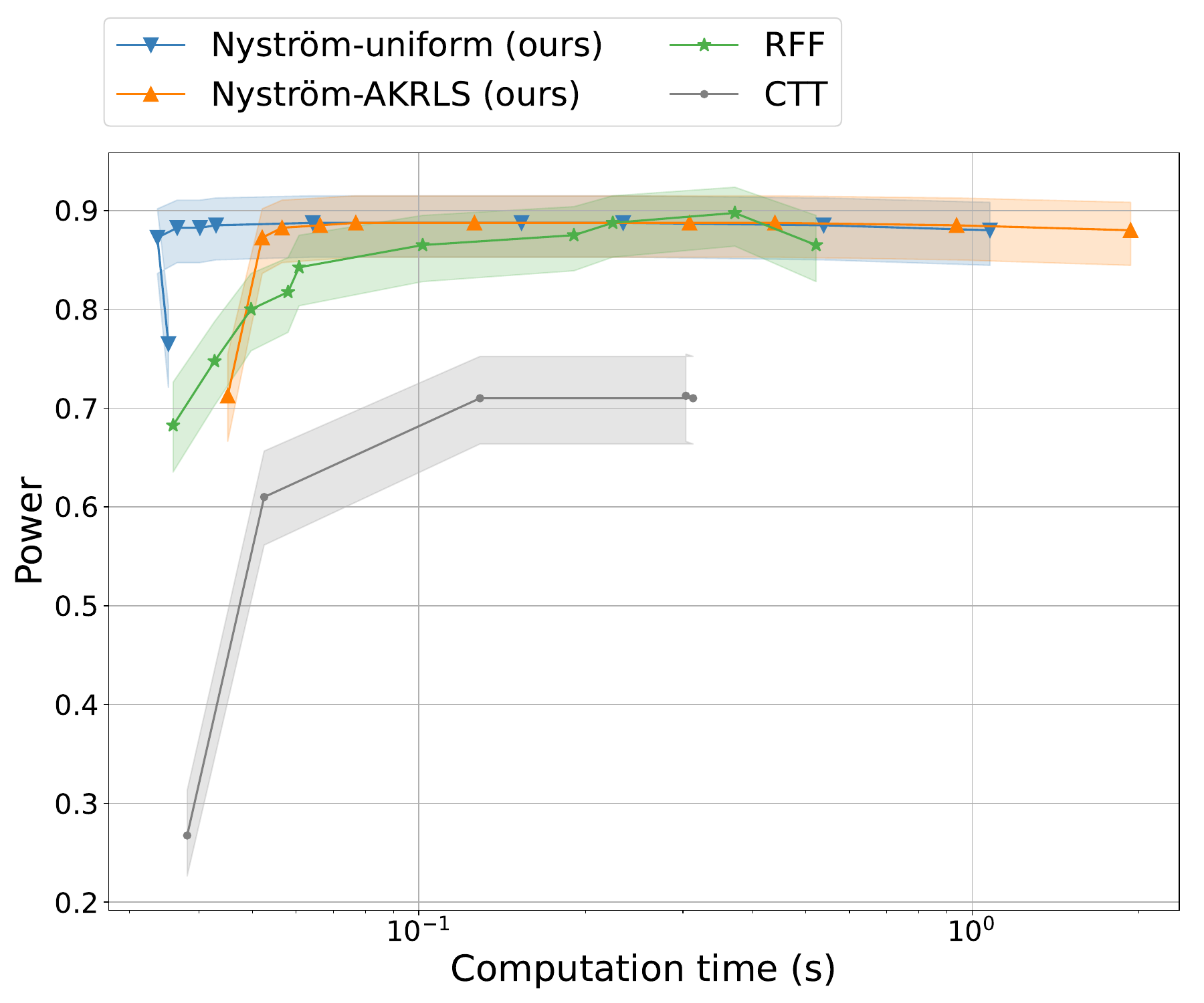}
		\label{fig:cg_pvst}
	}
	\hfill
	\subfigure
	{%
		\includegraphics[width=0.48\linewidth]{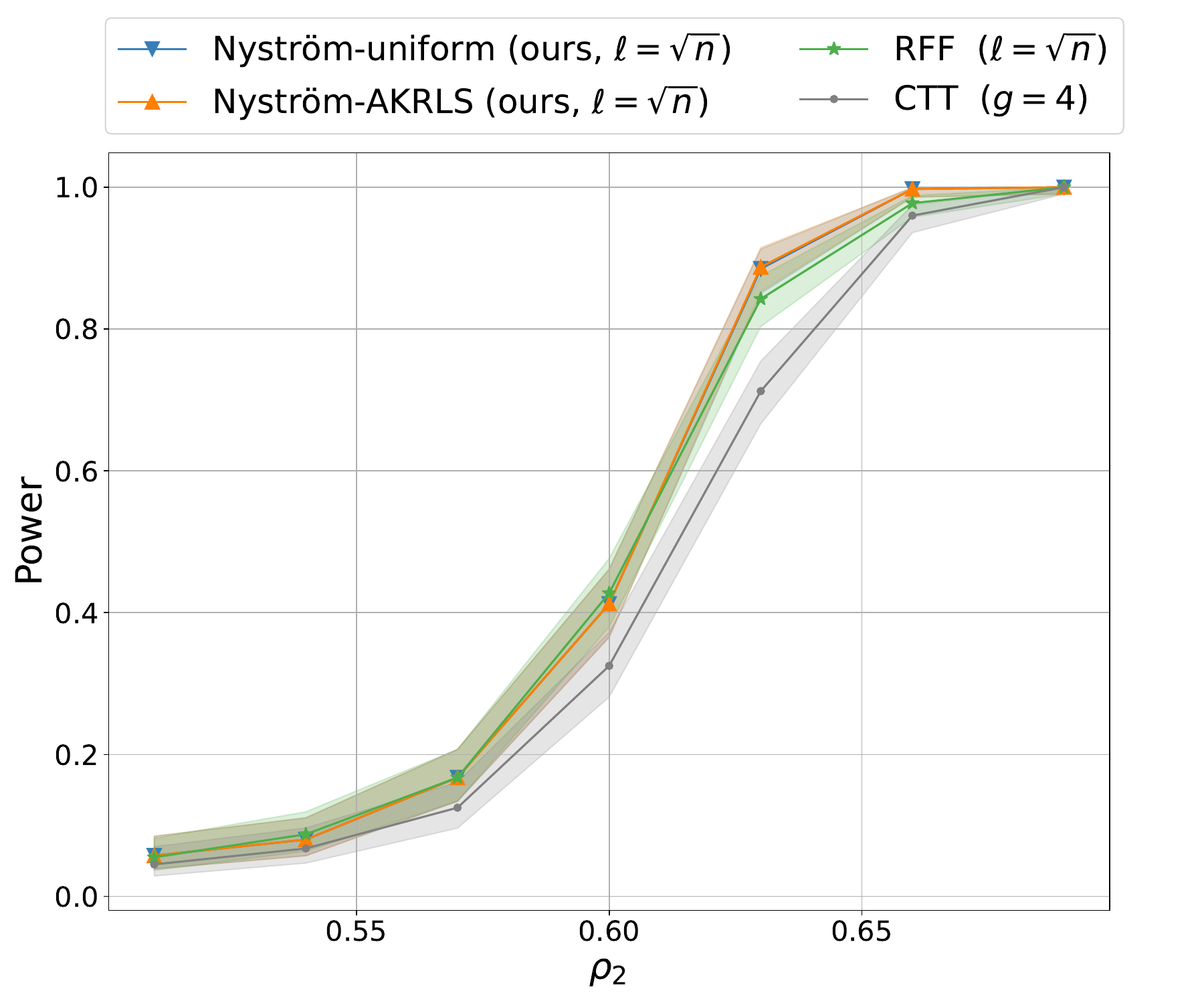}
		\label{fig:cg_pvsrho}
	}
	\caption{\textbf{Correlated Gaussians.} Left: power against computation time ($\rho_2=0.63$). The number of features $\ell$ is varied between 14 and 1500. We used CTT with compression levels $g=0,1,2,3,4$. Right: power against $ρ_2$.}
	\label{fig:cg_results}
\end{figure}

\begin{figure}
	\centering
	\subfigure
	{%
		\includegraphics[width=0.48\linewidth]{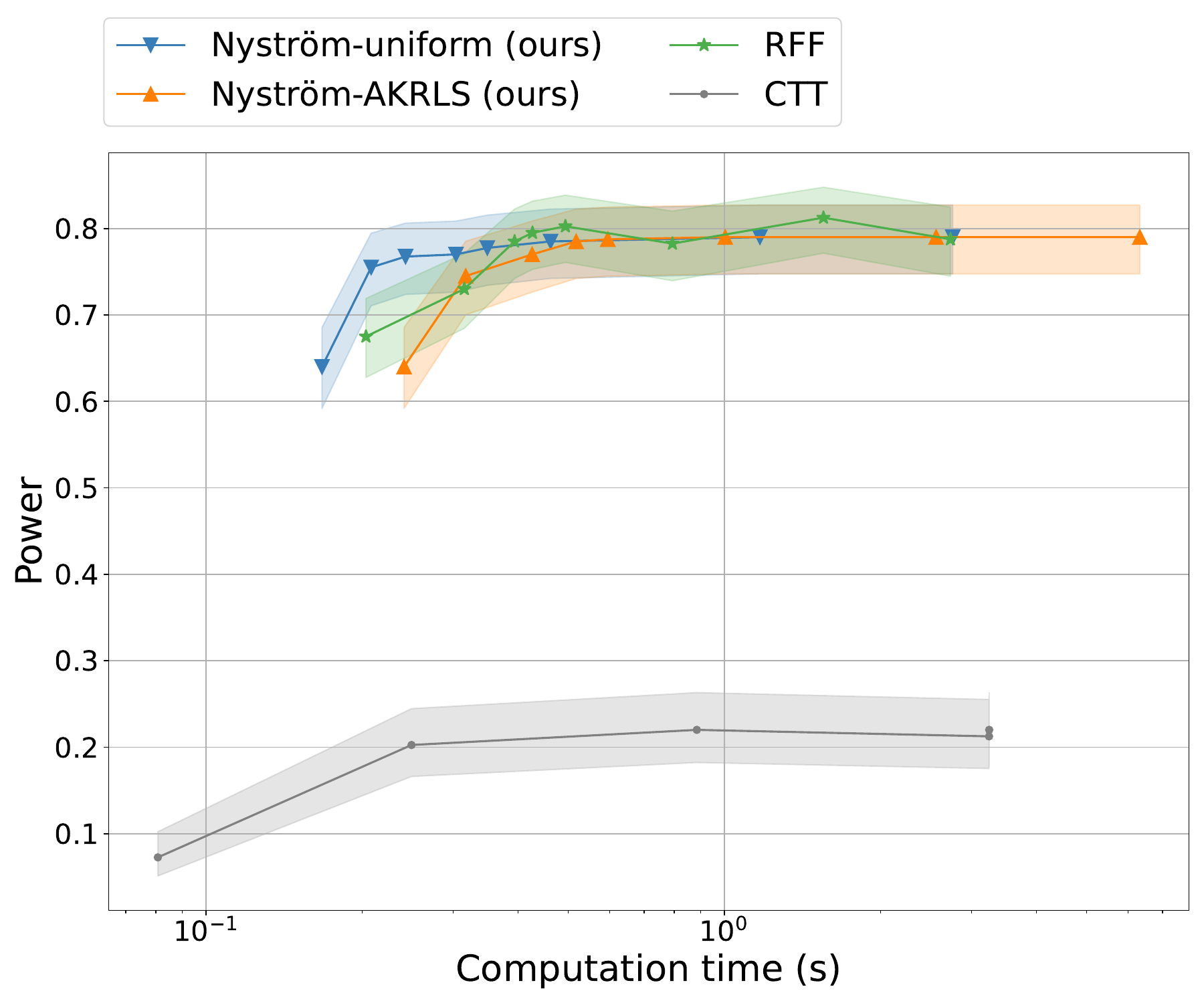}
		\label{fig:susy_pvst}
	}
	\hfill
	\subfigure
	{%
		\includegraphics[width=0.48\linewidth]{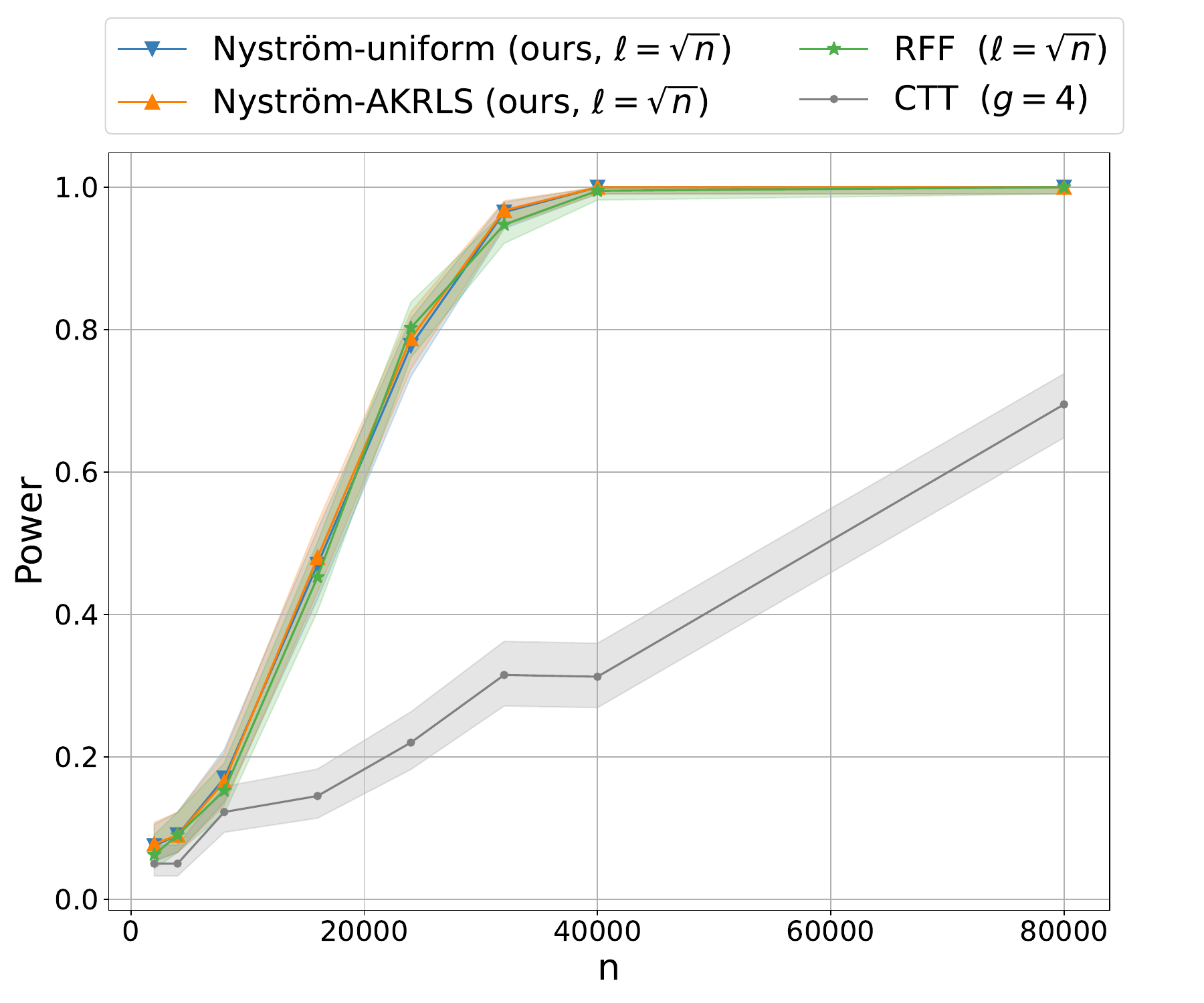}
		\label{fig:susy_pvsn}
	}
	\caption{\textbf{Susy.} Left: power against computation time ($n=24000$). The number of features $\ell$ is varied between 30 and 1540. We used CTT with compression levels $g=0,1,2,3,4$. Right: power against sample size.}
	\label{fig:susy_results}
\end{figure}

\begin{figure}
	\centering
	\subfigure
	{%
		\includegraphics[width=0.48\linewidth]{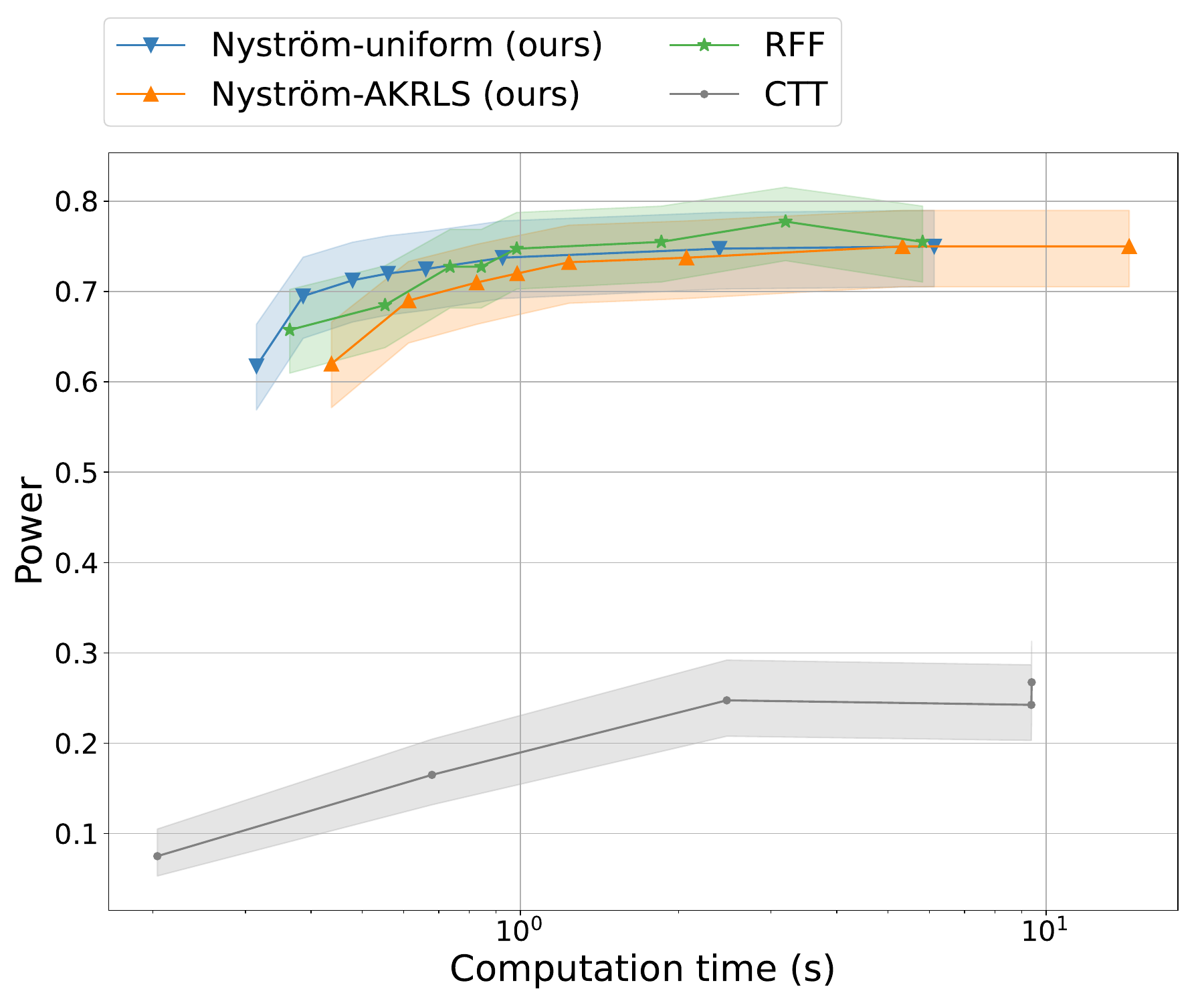}
		\label{fig:higgs_pvst}
	}
	\hfill
	\subfigure
	{%
		\includegraphics[width=0.48\linewidth]{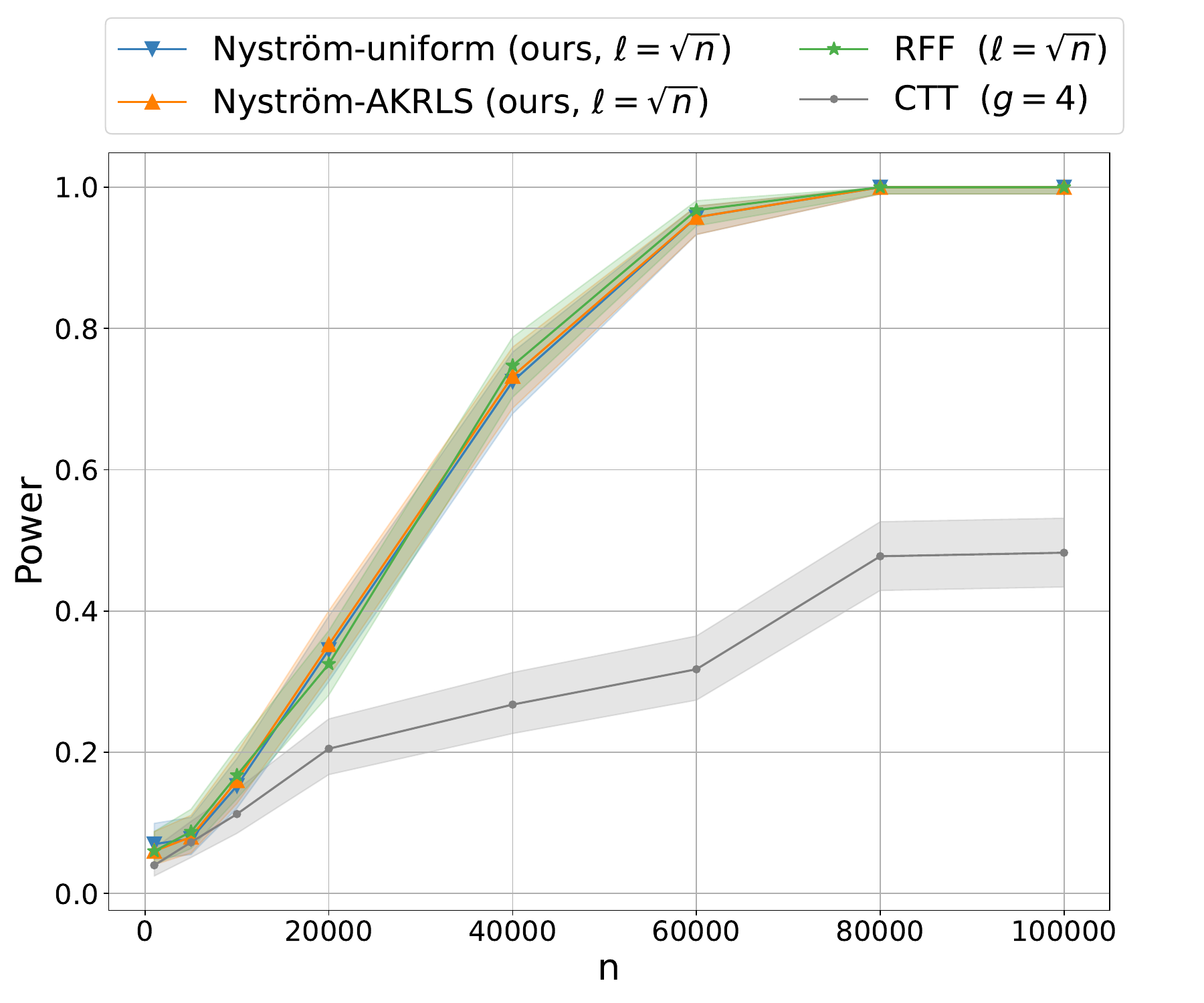}
		\label{fig:higgs_pvsn}
	}
	\caption{\textbf{Higgs.} Left: power against computation time ($n=40000$). The number of features $\ell$ is varied between 40 and 1000. We used CTT with compression levels $g=0,1,2,3,4$. Right: power against sample size.}
	\label{fig:higgs_results}
\end{figure}

In the left panels of \Cref{fig:cg_results,fig:susy_results,fig:higgs_results}, we report the power against the average computation time required to perform a full permutation test for the selected methods on the three datasets described in \Cref{s:datasets}.
We observe that our approach matches the results obtained using RFF, and outperforms CTT. The former are themselves known to empirically outperform other linear-time test statistics existing in the literature~\pcite{choi2024computational}. 
In the right panels of \Cref{fig:cg_results,fig:susy_results,fig:higgs_results}, we fix the number of features to $\nf=\sqrt{n}$ and compare the same methods while varying the parameter $ρ_2$ for the synthetic dataset and the number of samples $n$ for Susy and Higgs.

\begin{figure}
	\centering
	\subfigure[\cc{\textbf{Correlated Gaussians}.}]{
		\includegraphics[width=0.30\linewidth]{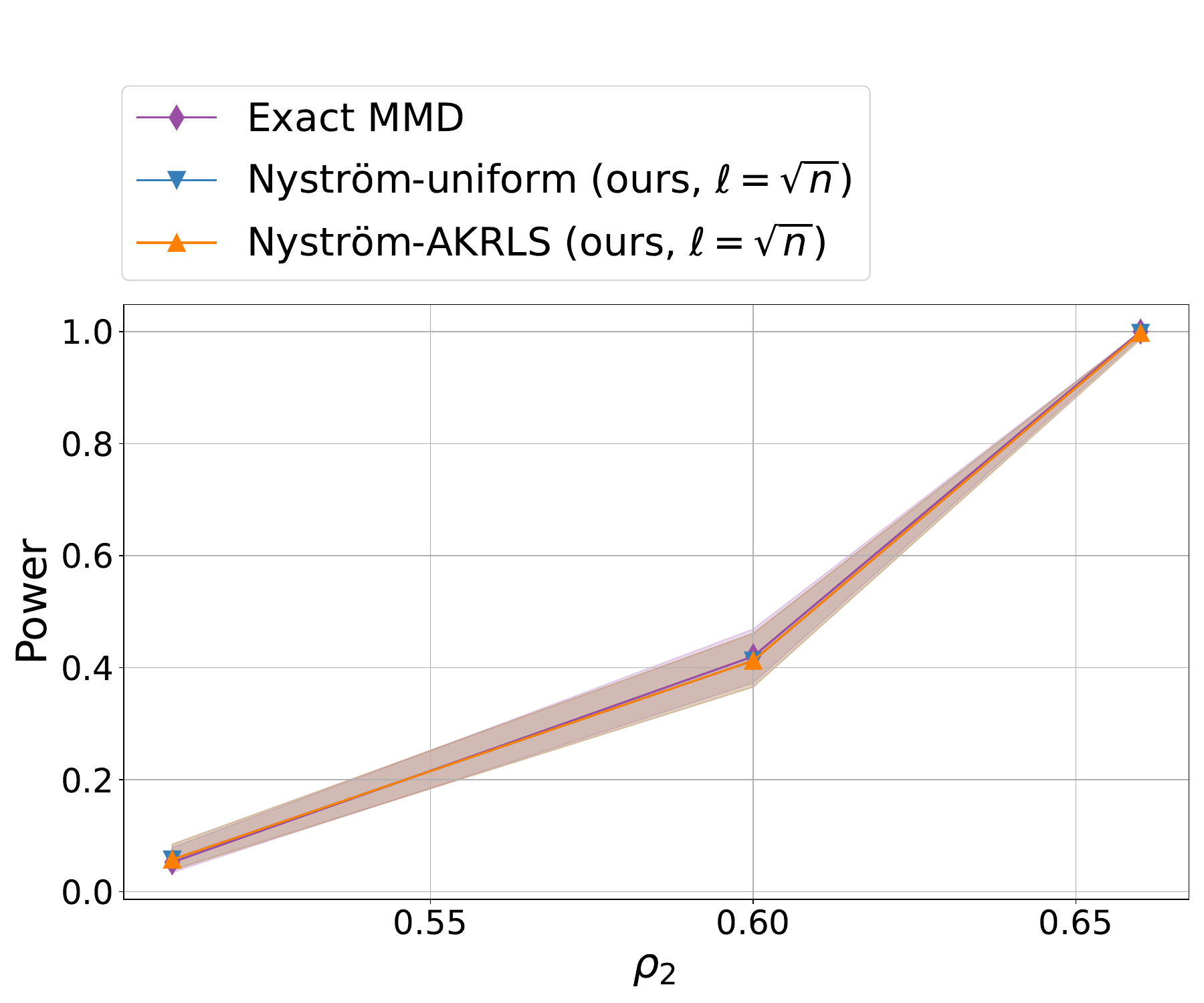}
		\label{fig:cg_pvsell_exact}
	}
	\subfigure[\cc{\textbf{Susy}.}]{
		\includegraphics[width=0.30\linewidth]{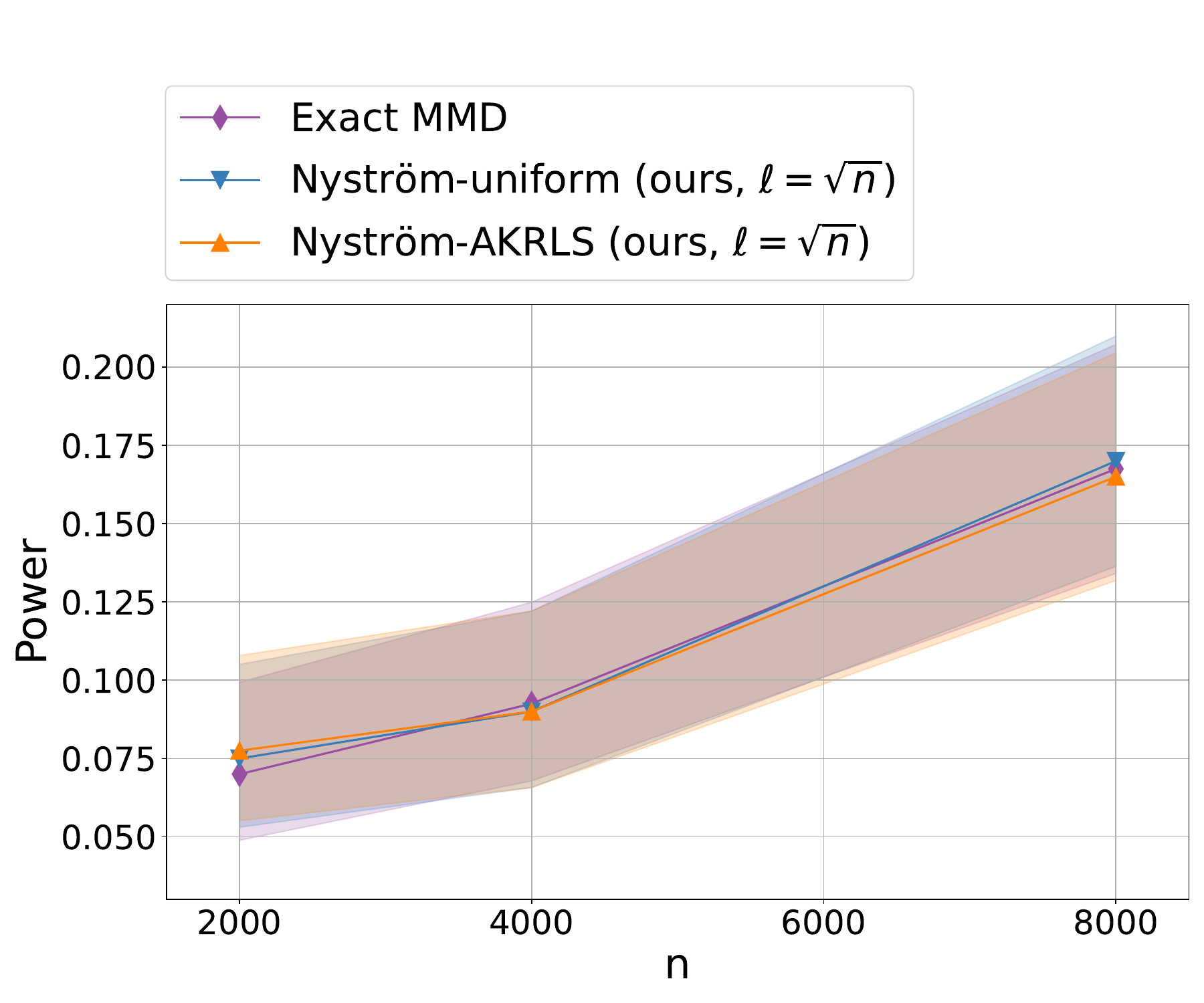}
		\label{fig:susy_pvsell_exact}
	}
	\subfigure[\cc{\textbf{Higgs}.}]{
		\includegraphics[width=0.30\linewidth]{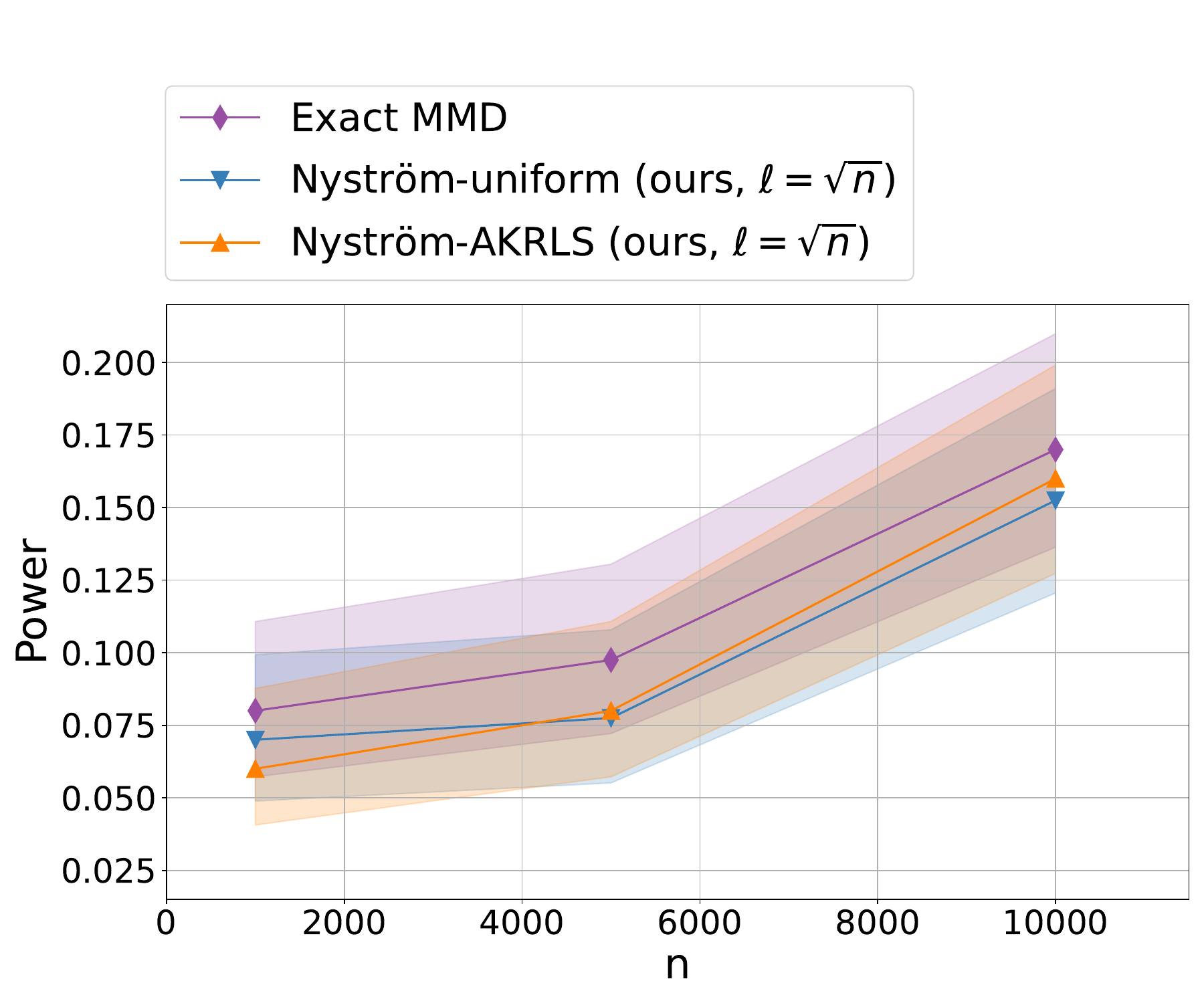}
		\label{fig:higgs_pvsell_exact}
	}
	\caption{Comparison with exact MMD.}
	\label{fig:comp_exact}
\end{figure}

We also report in \Cref{fig:comp_exact}  the results of the exact MMD estimators for smaller values of $n$, where computations remain feasible for Susy and Higgs, and for a small set of values of $\rho_2$ in the synthetic dataset. In these settings, one can clearly see that the efficiency of our approximate test procedure is obtained without compromising the power of the test.

\begin{figure}
	\centering
	\subfigure[\cc{\textbf{Correlated Gaussians}, $\rho_2=0.63$}]{
		\includegraphics[width=0.48\linewidth]{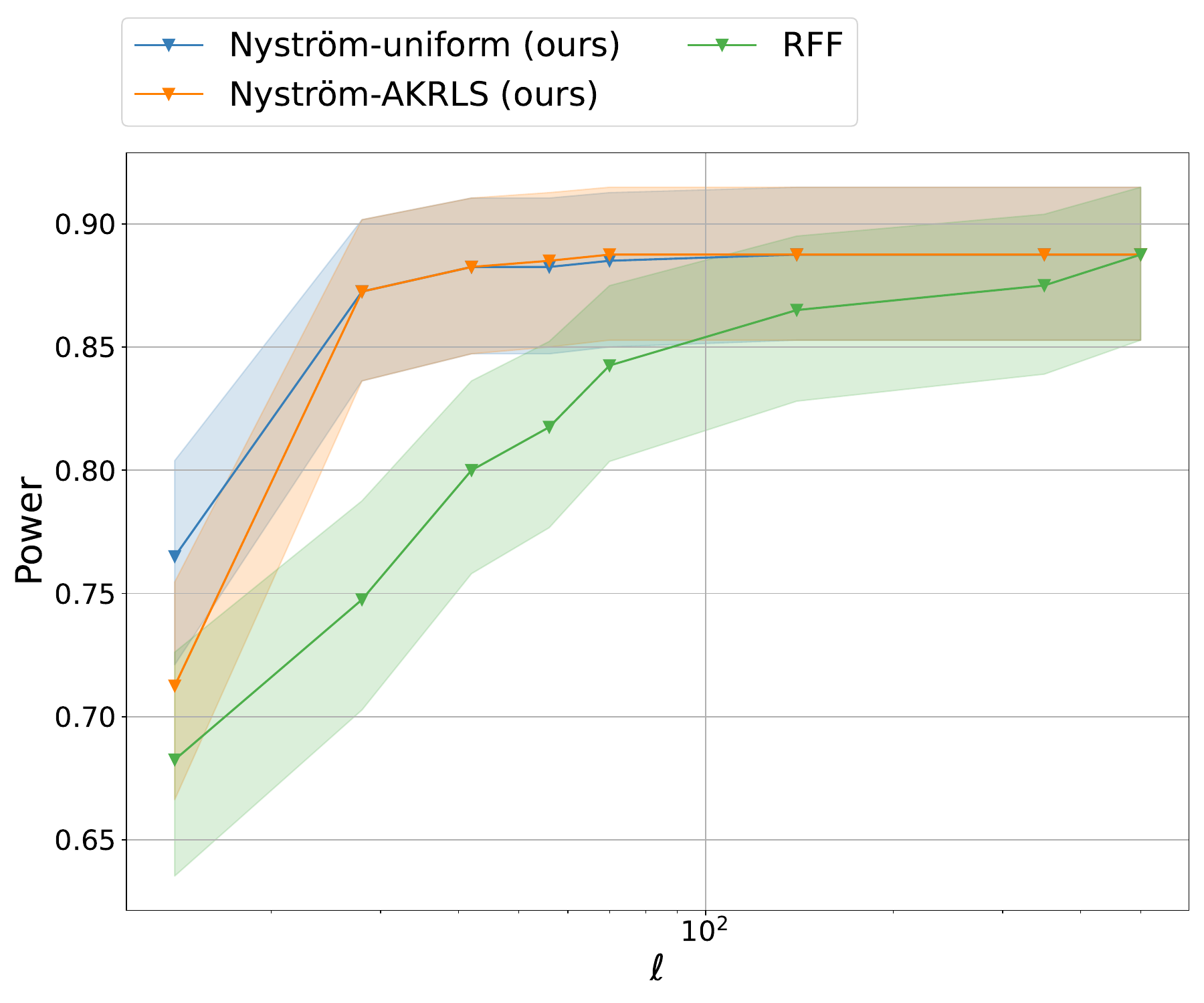}
		\label{fig:cg_pvsell}
	}
	\subfigure[\cc{\textbf{Susy},  $n=24000$}]{
		\includegraphics[width=0.48\linewidth]{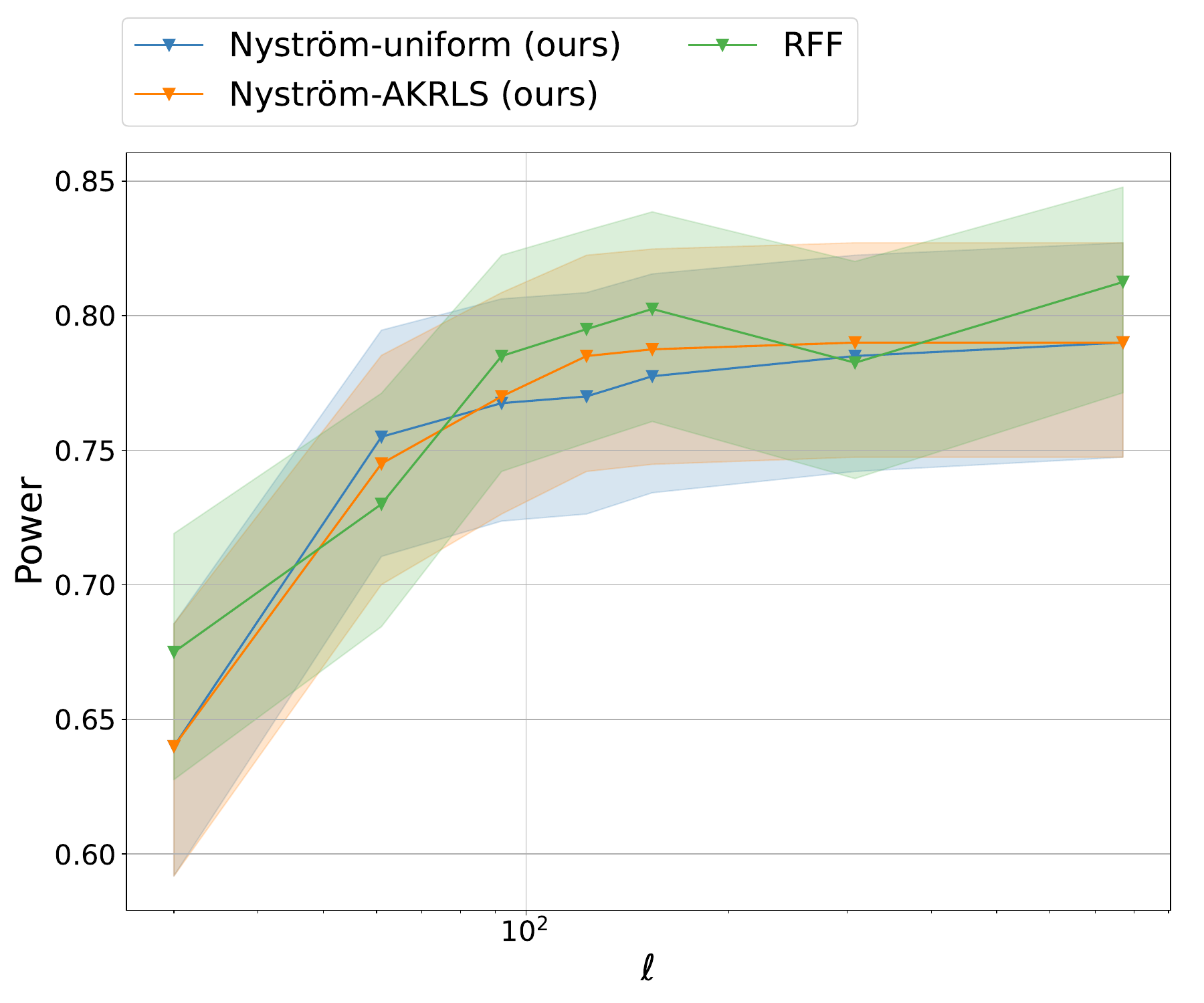}
		\label{fig:susy_pvsell}
	}
	\centering
	\subfigure[\cc{\textbf{Higgs}, $n=40000$}]{
		\includegraphics[width=0.48\linewidth]{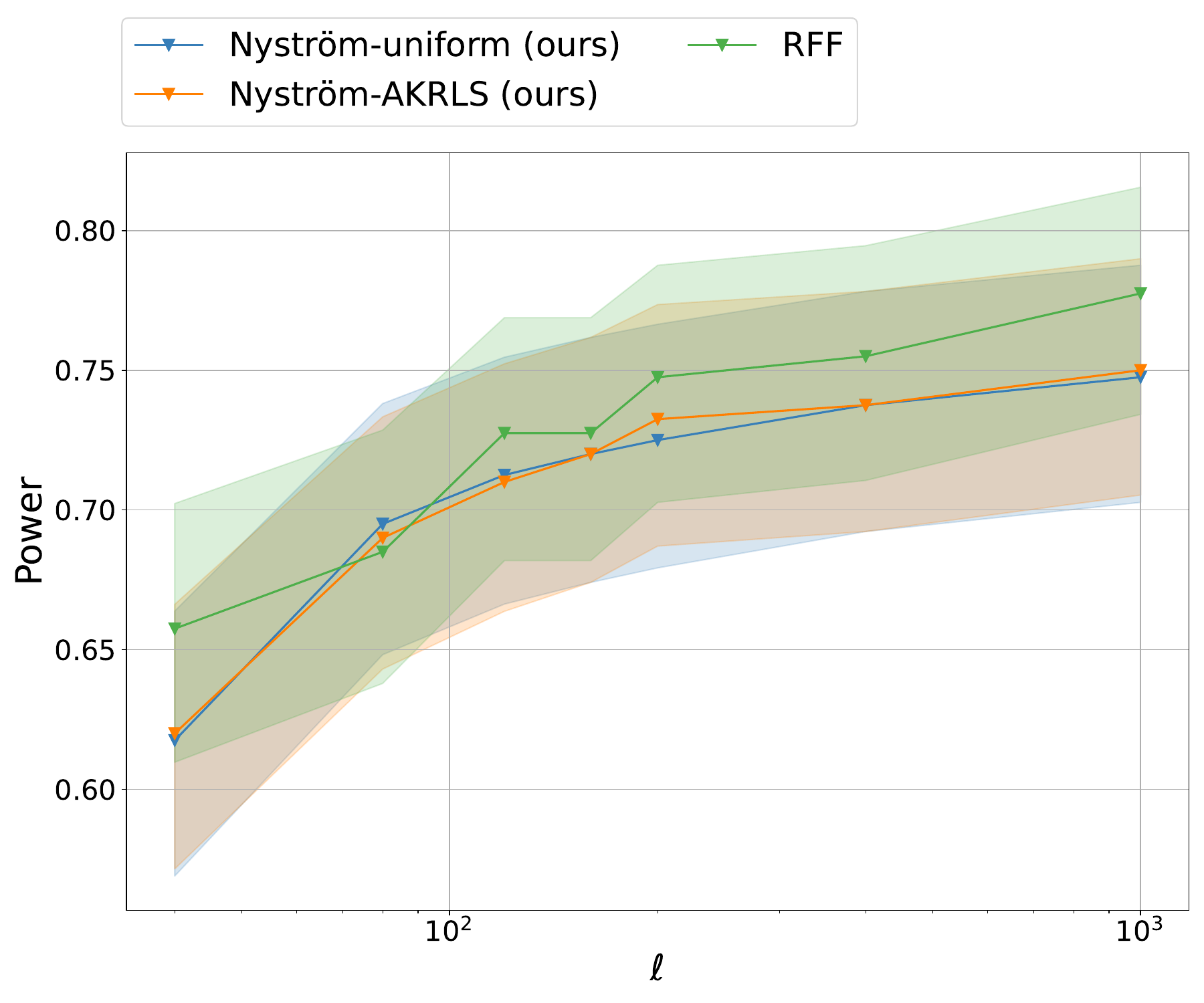}
		\label{fig:higgs_pvsell}
	}
	\caption{Power against number of features.}
	\label{fig:pvsell}
\end{figure}

In \Cref{fig:pvsell}, we report the power as a function of the number of features, which for our approach corresponds to the number of Nyström landmarks, and for RFF corresponds to the number of (real) features used in the approximation. Our approach performs overall similarly to RFF. 

\begin{figure}
	\centering
	\begin{minipage}{0.48\linewidth}
		\centering
		\includegraphics[width=\linewidth]{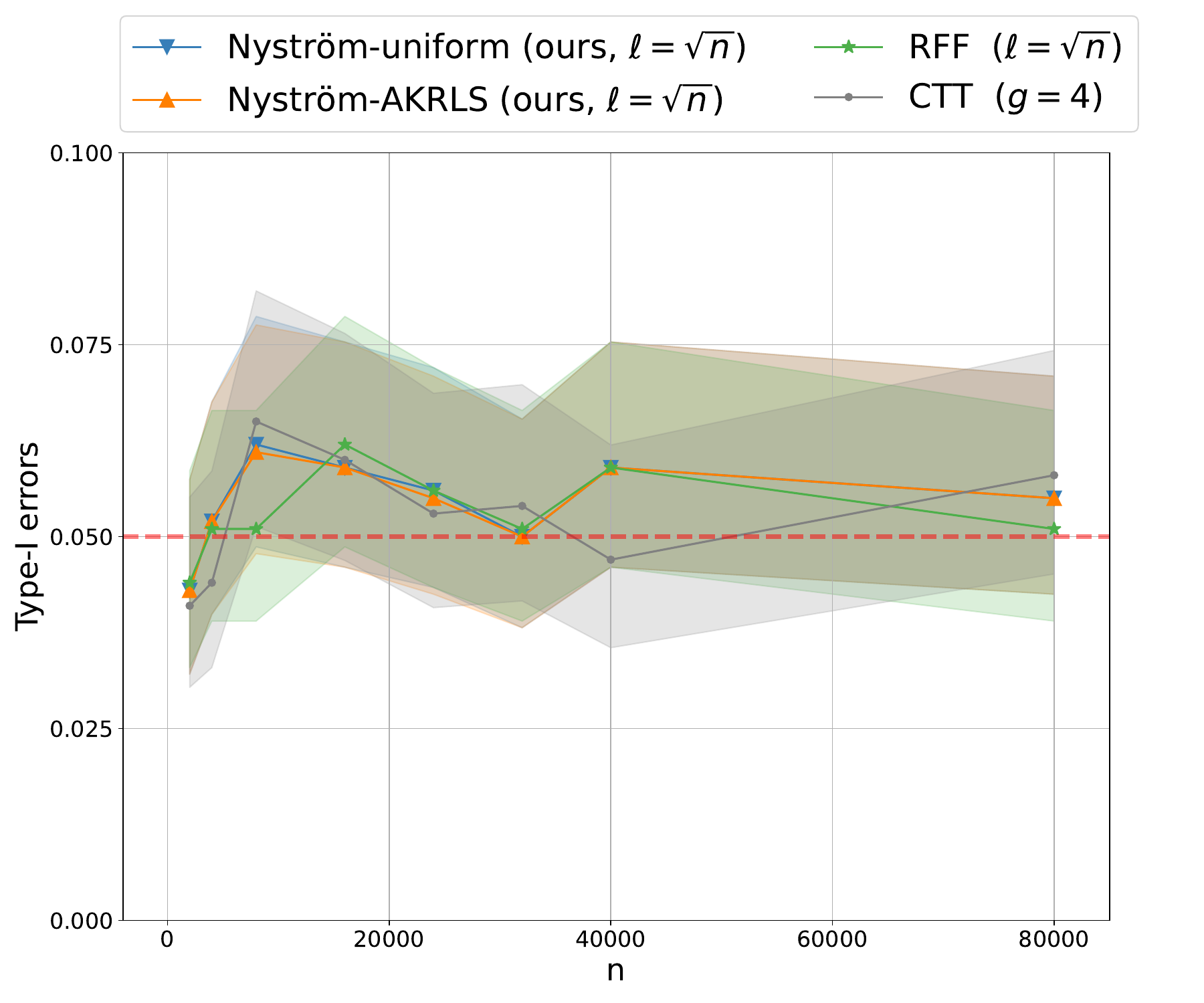}
		
		\small \cc{\textbf{Susy}, $n=24000$}
		\label{fig:susy_fpr}
	\end{minipage}
	\hfill
	\begin{minipage}{0.48\linewidth}
		\centering
		\includegraphics[width=\linewidth]{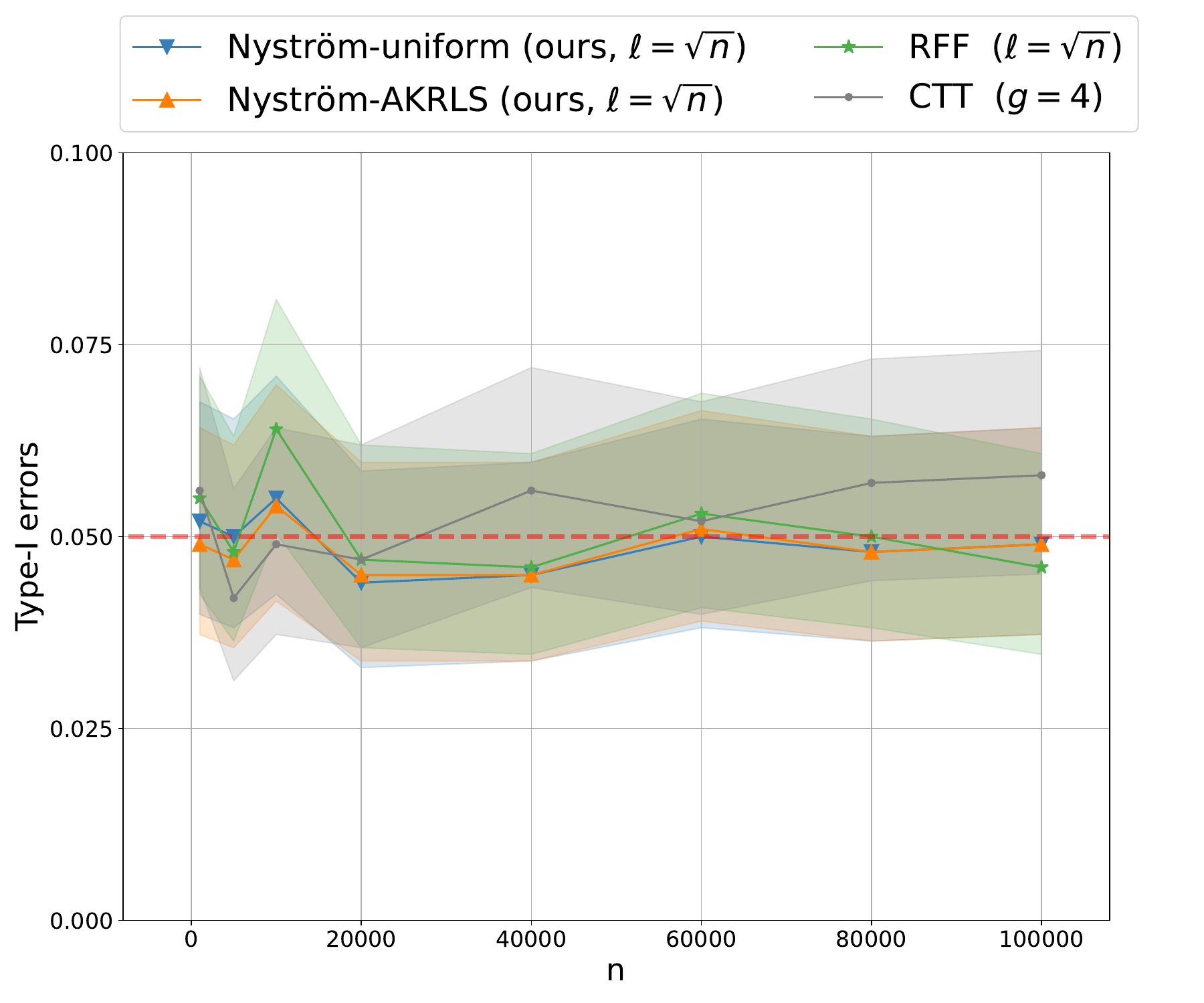}
		
		\small \cc{\textbf{Higgs}, $n=40000$}
		\label{fig:higgs_fpr}
	\end{minipage}
	\caption{Rate of type-I errors against sample size. The number of features $\ell$ for our methods and RFF is varied between 40 and 1000 for the Higgs dataset and between 30 and 770 for the Susy dataset. We used CTT with compression levels $g=0,1,2,3,4$.}
	\label{fig:fpr}
\end{figure}

In \Cref{table:fpr_cg} and \Cref{fig:fpr}, we show the rate of type-I errors (with a level $\alpha=0.05$) for the synthetic dataset and Susy and Higgs datasets (at varying sample size).

\renewcommand{\arraystretch}{1.2}
\begin{table*}
\begin{center}
\caption{Type-I error rates for the synthetic dataset ($n=5000$).}
\label{table:fpr_cg}
\begin{tabular}{l|c}
\toprule
 Method & Type-I error rates \\
 \midrule
 Nyström-uniform (ours)  & $0.044_{-0.011}^{+0.015}$ \\
 Nyström-AKRL (ours)  & $0.044_{-0.011}^{+0.015}$ \\
 RFF  & $0.049_{-0.012}^{+0.015}$ \\
 CTT  & $0.044_{-0.012}^{+0.013}$\\
 \bottomrule
\end{tabular}
\end{center}
\end{table*}

\section{Conclusion}

We introduced a computationally efficient procedure for two-sample testing based on a data-adaptive Nyström approximation of the maximum mean discrepancy. We established non-asymptotic power guarantees for the resulting test and showed that it achieves minimax optimal separation rates in terms of maximum mean discrepancy. The procedure is simple to implement, and performs favorably compared to existing state-of-the-art methods, \new{both in practice and in terms of theoretical tradeoff between statistical power and computational resources.}
Future work includes extending the proposed test statistic to support aggregated testing procedures and the handling of weighted samples.


\ifejs\begin{acks}[Acknowledgments]\else\section*{Acknowledgments}\fi
	The authors would like to thank Lester Mackey and Carles Domingo-Enrich for their help and feedback regarding the compress-then-test method.
	\ifejs\end{acks}\else\fi
	
	\ifejs\begin{funding}\else\section*{Funding}\fi
	All the authors acknowledge the financial support of the European Research Council (grant SLING 819789). L. R. acknowledges the financial support of the European Commission (Horizon Europe grant ELIAS 101120237), the US Air Force Office of Scientific Research (FA8655-22-1-7034), and the Ministry of Education, University and Research (FARE grant ML4IP R205T7J2KP; grant BAC FAIR PE00000013 funded by the EU - NGEU). This work represents only the view of the authors. The European Commission and the other organizations are not responsible for any use that may be made of the information it contains.									
	\ifejs\end{funding}\else\fi

	\ifejs
	\else
		\clearpage
		\printbibliography
	\fi

\ifejs
	\begin{appendix}%
\else
	\newpage\appendix%
\fi

\section{Additional notations}

For any projector $P$, we denote $P^⟂=I-P$. For an event $B \in \mathcal{B}$ we denote by $B^c$ the complement of the event $B$, \ie $B^c \coloneqq E \setminus B$ \new{where $E$ denotes the sample space}. 

\new{The notation $\n{·}$ is used both for the $l_2$ norm of vectors and for the operator norm of operators. We recall that the RKHS norm is denoted $\nrkhs{·}$.}

\section{Proof for the level (\Cref{r:level_hemerik_th2})}
\label{s:proof_level}

\NewDocumentCommand\Tipts{g}{\hat{Ψ}\IfNoValueF{#1}{_{#1}}} 
\NewDocumentCommand\Toipts{gO{Z}O{Σ}O{W}}{\prt*{\hat{Ψ}_{#2}^{#3}(#4)}\IfValueT{#1}{_{(#1)}} } 


For clarity, in this section we make explicit the dependence of the permuted test statistics~$\oipts{i}$ on the pooled dataset~$\XY$,  the landmarks~$\ldms$ and the set of permutations~$Σ\de(σ_0\de \textrm{id}, σ_1,…, σ_{\np})$. Accordingly, we write $\Toipts{j}$ in place of $\oipts{j}$, $0≤j≤\np$.

\begin{tproofof*}{r:level_hemerik_th2}{} 
	We consider $\np \geq 1$ permutations $\sigma_1, \dots, \sigma_\np$, drawn independently and identically from the uniform distribution over $\new{S_n}$, the set of permutations of $\{1, \dots, n\}$. 
	Let 
	\begin{align*}
		\Toipts{0}≤…≤\Toipts{\np}
	\end{align*}
	denote the ordered test statistics corresponding to permutations in $Σ$. For notational clarity and consistency, we write the unpermuted test statistic as $\ts[\W]$, instead of $\ts$.
	
	It follows from the law of total probability and from definition of Algorithm~\ref{al:main_test} that
	\begin{align*}
		\P[\text{reject}]
		&= \P*[\ts[W]  > \Toipts{b_α}] \\
			&\quad + \P*[\text{reject}|\ts[W]  = \Toipts{b_α}]·\P*[\ts[W]  = \Toipts{b_α}] \\
		&= \E\brk*{\mathds{1}\prt*{\ts[W]  > \Toipts{b_α}}
			+ p(\W,\ldms,Σ)\,  \mathds{1}\prt*{\ts[W] = \Toipts{b_α}}},
	\end{align*}
	where $p(\W,\ldms,Σ) \de \frac{\alpha(\np+1) - \tsl^>}{\tsl^=}$ denotes the randomized rejection probability used in \Cref{al:main_test} that depends on 
	\begin{align*}
		\tsl^> \de \tsl^>(\W,\ldms,Σ) \de \#\cb*{0 ≤ b ≤ \np : \Tipts{b} > \Toipts{b_α}}
	\end{align*}
	and 
	\begin{align*}
		\tsl^= \de \tsl^=(\W,\ldms,Σ) \de \#\cb*{0 \leq b ≤ \np : \Tipts{b} = \Toipts{b_α}}.
	\end{align*}
		
	Observing that $Σ$ and $Σσ_j^{-1}$ have the same distribution (as unordered sets) for any $j∊\irange{\np}$, letting $j\sim \cU(\irange[0]{\np})$ and $h=σ_j$, it holds that $Σ$ and $Σh^{-1}$ have the same distribution as well.
	
	As a consequence, denoting $π(W)=(π_i(W))_{j=1}^n$ the probabilities according to which the landmarks are sampled, the above probability can be expressed as
	\begin{align*}
		\P[\text{reject}] 
		&= \E_{Σ,h,\W} \E_{\ldms_k=W_{i_k}, i_k\sim π(W) }
			\Big[\mathds{1}\prt*{\ts[W]  > \Toipts{b_α}[Z][Σh^{-1}][\W]}
			\\&\quad
			+ p(\W,\ldms,Σh^{-1})\, \mathds{1}\prt*{\ts[W] = \Toipts{b_α}[Z][Σh^{-1}][\W]}\Big].
	\end{align*}
	Furthermore, since the samples in $W$ are exchangeable under the null, we have $hW \stackrel{(d)}{=} W$ and thus
		\begin{align*}
		\P[\text{reject}] &= \E_{Σ,h,\W} \E_{\ldms_k=W_{i_k}, i_k\sim π(W) }
		\Big[\mathds{1}\prt*{\ts[h W]  > \Toipts{b_α}[Z][Σh^{-1}][h \W]}
		\\&\quad
		+ p(h\W,\ldms,Σh^{-1})\, \mathds{1}\prt*{\ts[h W] = \Toipts{b_α}[Z][Σh^{-1}][h \W]} \Big].
	\end{align*}

	Observing that $(Σh^{-1})(h \W) = Σ(\W)$, it holds $p(h\W,\ldms,Σh^{-1})=p(\W,\ldms,Σ)=:p$ and
	\begin{align*}
		\P[\text{reject}] 
		&= \E \E_{\substack{\ldms_k=(hW)_{i_k}\\ i_k\sim π(hW) }}
			\Big[\mathds{1}\prt*{\ts[h W]  > \Toipts{b_α}}
			\\&\quad
			+ p(W,\ldms,Σ)\, \mathds{1}\prt*{\ts[h W] = \Toipts{b_α}}\Big].
	\end{align*}

	For landmark sampling strategies $\pi$ that are equivariant under permutations of the pooled dataset---such as uniform and AKRLS sampling---we have $π(h\W)=hπ(\W)$. As a consequence, using the fact that  $(hW)_{i_k} = W_{h^{-1}(i_k)}$ and $i_k \sim h\pi(W) \iff h^{-1}i_k \sim \pi(W)$, we obtain
	\begin{align*}
	\P[\text{reject}] 
		&= \E \E_{\substack{\ldms_k=(W)_{h^{-1}i_k}\\ i_k\sim h π(W) }}
			\Big[ \mathds{1}\prt*{\ts[h W]  > \Toipts{b_α}}
			\\&\quad\quad
			+ p(\W,\ldms,Σ)\, \mathds{1}\prt*{\ts[h W] = \Toipts{b_α}} \Big] \\
		&= \E \E_{\substack{\ldms_k=(W)_{i_k}\\ i_k\sim π(W) }}
			\Big[\mathds{1}\prt*{\ts[h W]  > \Toipts{b_α}}
			\\&\quad\quad
			+ p(\W,\ldms,Σ)\, \mathds{1}\prt*{\ts[h W] = \Toipts{b_α}}\Big].
	\end{align*}
	
	Finally, using the definition of $h$, we can write both expectations over $h$ as 
	\begin{align*}
		\P[\text{reject}] 
		 &= \E_{Σ,\W,\ldms}
			\Bigg[
			(\np+1)^{-1} \Bigg(
				\#\cb*{0≤j≤\np : \mathds{1}\prt*{\Toipts{j}[Z][Σ][\W]  > \Toipts{b_α}[Z][Σ][\W]}}
				\\&\quad\quad + p(\W,\ldms,Σ)\, \#\cb*{0≤j≤\np : \mathds{1}\prt*{\Toipts{j}[Z][Σ][\W]  = \Toipts{b_α}[Z][Σ][\W]}}
				\Bigg)
			\Bigg] \\
		 &= \E_{Σ,\W,\ldms}
			\brk*{ (\np+1)^{-1} \prt*{ \tsl^>(\W,\ldms,Σ) + \prt*{ \frac{\alpha(\np+1) - \tsl^>(\W,\ldms,Σ)}{\tsl^=(\W,\ldms,Σ)} }\, \tsl^=(\W,\ldms,Σ)} } \\
		 &= α,
	\end{align*}
	where the last inequality follows from the definition of $p$.
\end{tproofof*}

\color{black}

\section{MMD Approximation Quality: Additional Results and Proofs}
\label{app:quality_mmd_approx_new}

We first recall some concentration inequalities in \Cref{s:concentration_inequalities}, then introduce in \Cref{s:mmd_additional_lemmas} some additional lemmas which we use in \Cref{s:proof_bound_MMD_nys_new} in order to prove \Cref{r:bound_MMD_nys_new}.

We recall that $\covM\de \frac{\nX}{n} \covP + \frac{\nY}{n} \covQ$. 
In this section, we use the additional notations $\ecovM\de \frac{\nX}{n} \ecovX + \frac{\nY}{n} \ecovY$, $\recovM\de \ecovM+λI$ and $\rcovM\de \covM+λI$. 

\subsection{Concentration inequalities}
\label{s:concentration_inequalities}

The following lemma is a restatement of \tcite[Theorem 2.10]{boucheron2013ConcentrationInequalitiesNonasymptotic}.
\begin{tlemma}{Bernstein's inequality for real-valued random variables}{real_bernstein_var}
	Let $X_1,…,X_n$ be independent real-valued random variables. Assume that there exist $c>0,v>0$ such that $|X_i|≤c$ almost everywhere and $\sum_{i=1}^n \E[X_i²] ≤ v$. 
	Then for any $δ∊]0,1[$ it holds
	\begin{align*}
		\P*[\absv*{\sum_{i=1}^n X_i - \sum_{i=1}^n\E[X_i]} ≥ \sqrt{2v\log(2/δ)}+\frac{c\log(2/δ)}{3}]
		&≤ δ.
	\end{align*}
\end{tlemma}
\begin{tproofof*}{r:real_bernstein_var}
This is a direct corollary of \pcite[Theorem 2.10]{boucheron2013ConcentrationInequalitiesNonasymptotic}. 
Indeed if $\lvert X_i \rvert ≤c'$ almost surely, then $|\max(X_i,0)|≤c'$ almost surely. Moreover, for any $q≥3$:
\begin{align*}
	\sum_{i=1}^n \E[\max(X_i,0)^q] 
	&≤ \sum_{i=1}^n \E[X_i²|X_i|^{q-2}] \\
	&≤ c'^{q-2} v 
	 = 3^{q-2}  (c'/3)^{q-2} v 
	 ≤ \frac{q!}{2} (c'/3)^{q-2} v.
\end{align*}
The claimed result follows from \pcite[Theorem 2.10]{boucheron2013ConcentrationInequalitiesNonasymptotic} with constants $v$ and $c\de c'/3$.
\end{tproofof*}

\begin{tlemma}{{{{\tcite[Theorem 3.5]{pinelis1994optimum}}}}}{concentration_iid_hilbert}
	Let $X_1, \dots, X_n$ be \tiid \new{zero-mean} random variables on a separable Hilbert space $(\cX,\norm{\cdot})$ such that $\sup_{i=1, \dots, n}\norm{X_i}\leq A$ almost surely, for some real number $A > 0$. Then, for any $δ \in (0, 1)$, it holds with probability at least $1-δ$ that
	\[
	\norm*{\frac{1}{n}\sum_{i=1}^n X_i } 
	\leq A\frac{\sqrt{2\log(2/δ)}}{\sqrt{n}}.
	\]
\end{tlemma}

\new{%
We also restate \tcite[Proposition 9]{rudi2015LessMoreNystrom}, which generalizes \pcite[Theorem 2]{alaoui2015FastRandomizedKernel}.}
\begin{tlemma}{{{{\tcite[Proposition 9]{rudi2015LessMoreNystrom}}}}}{prop9_rudi}
\new{%
	Let $n,m$ be positive integers. Consider $A∊ℝ^{n×n}$ and denote by $a_i$ the $i$-th column of $A$.
	Let $m≤n$ and $I=\cb{i_1,…,i_m}$ be a subset of $\cb{1,…,n}$ formed by $m$ elements chosen randomly with replacement, according to a distribution that associates the probability $p_i$ to the element $i∊\cb{1,…,n}$. 
	Assume that there exists a $β∊(0,1]$ such that the probabilities $p_1,…,p_n$ satisfy $p_i≥β \frac{‖a_i‖²}{\Tr(AAᵀ)}$ for all $i$. For any $δ>0$ it holds with probability $1-δ$:
	\begin{align*}
		\n*{AAᵀ-\frac{1}{m} \sum_{i∊I} a_ia_iᵀ} 
		&≤ \frac{2‖A‖²\log(\tfrac{2n}{δ})}{3m} + \sqrt{\frac{2‖A‖²\Tr(AAᵀ)\log(\tfrac{2n}{δ} )}{β m} }.
	\end{align*}
}
\end{tlemma}

\subsection{Additional lemmas}
\label{s:mmd_additional_lemmas}

The next lemma quantifies the error induced by a leverage scores-based Nyström approximation, in operator norm, when the sampling is done from the pooled dataset $\W$.

\begin{tlemma}{}{bound_proj_halfcov_new}
	Consider samples $X_1, …, X_{\nX} \diid P$ and $Y_1, …, Y_{\nY} \diid Q$.
	Let $δ>0$, $z≥1$ and $λ_0>0$.  Consider the $(z,λ_0,δ/3)$-Approximate Kernel Ridge Leverage Scores $(\als{i})_{1≤i≤n}$  
	computed on the pooled dataset $W = \{X_1, \dots, X_{\nX}, Y_1, \dots, Y_{\nY}\}$.

	Let $λ>0$ and define the probability distribution $\pals$  on $\irange{n}$ defined by $\pals(i)\de\als{i}/(\sum_{i=1}^n \als{i})$. %
	Let $\cI=\cb{i_1,…,i_{\nf}}$ be a multiset of $\nf \geq 1$ indices drawn independently with replacement from $\pals$. 
	Let $\Pm$ be the orthogonal projection onto $\Hm\de\spa\cb{\fmap{x_j}|j∊\cI}$.
	Then under \Cref{a:bounded_kernel,a:poly_decay_mean}, provided that
	\begin{itemize}
		\item $\nf ≥ 4 (1 + 2.4 z²\mdeff) \log(\tfrac{6n}{δ})$
		where $\mdeff\de \Tr(\covM(\covM+λI)^{-1})$
		\item $\lambda_0 ≤λ$. 
		\item $\frac{16\supk\log(2/δ)}{n}≤λ≤\noprkhs{\covM}$.  
	\end{itemize}
	it holds with probability at least $1-δ$ 
	\begin{align*}
		\noprkhs{\Pmo (\ecovM+λI)^{1/2}}² &≤ 3λ
		\quad\text{ where }\quad
		\ecovM\de \frac{\nX}{n} \ecovX + \frac{\nY}{n} \ecovY.
	\end{align*}
\end{tlemma}

Note that the assumptions are expressed as a function of the “mixed” effective dimension $\mdeff$, however the later can be related the effective dimensions \wrt $P$ and $Q$, as explained in the comment after Assumption~\ref{a:poly_decay_mean}

The proof is similar in spirit to \pcite[Lemma 7]{rudi2015LessMoreNystrom}, with two main differences: (1) the bound involves the empirical covariance and not the population version, which is sufficient in our setting and simplifies a few things and (2) the data is not \tiid given that we have samples from both~$P$ and~$Q$.
\newcommand\scv{\ftn H \ftn^*}

\begin{tproofof*}{r:bound_proj_halfcov_new}
	\proofparagraph{Step 1: Operator-theoretic bounds.}
	Consider the linear operator $\ftn:ℝ^n→\rkhs$, $\ftn w \de \sum_{i=1}^n w_i \fmap{W_i}$, with adjoint $\ftn^*h = [h(W_i)]_{i=1}^n$ for any $h \in \rkhs$. Note that $\frac{1}{n} \ftn\ftn^*=\ecovM$. 
	Let \( q_i \) denote the number of times index \( i \in \{1, \dots, n\} \) appears in the multiset \( \mathcal{I} \); that is,
	\[
	q_i \coloneqq \#\{s \in \{1, \dots, \l\} : i_s = i\}.
	\]
	
	Let $H$ be the diagonal matrix with entries $H_{ii}=\tfrac{q_i}{n \nf \pals{i}}$ when $\pals{i}≠0$, $H_{ii}=0$ otherwise. 
	
	Note that $\E[H_{ii}]=\tfrac{1}{n \nf\pals{i}} \E[\#\cb{j∊\irange{\nf}|i_j=i}]=\tfrac{1}{n}$, hence $\E[\scv | \W]=\ecovM$. 
	
	Observe that 
	\begin{align*}
		\rg(\ftn H \ftn^*)=\rg(\ftn H^{1/2})= \rg(\ftldms)=\rg(\Pm)
	\end{align*}
	since the operators involved are all finite-rank. 
	Indeed, the range of a linear map equals its closure when it is of finite dimension, and every operator above spans the finite-dimensional subspace spanned by the features of the landmarks.
		
Hence, by successively applying \tcite[Proposition 3 and Proposition 7]{rudi2015LessMoreNystrom}, we obtain that, almost surely,
	\begin{align}
		\noprkhs{\Pmo (\ecovM+λI)^{1/2}}²
		&≤ λ \noprkhs{(\scv+λI)^{-1/2} (\ecovM+λI)^{1/2}}²	
		 ≤ \tfrac{λ}{ 1-β(λ) }	
		 \label{e:bypropseven}
	\end{align}
	whenever  
	\begin{align*}
		β(λ) &\de λ_{max}\prt*{(\ecovM+λI)^{-1/2}(\ecovM-\scv)(\ecovM+λI)^{-1/2}} < 1.
	\end{align*}
	
	\proofparagraph{Step 2: Deterministic bounds on $\beta(\lambda)$.}
	
	Let us now control the term $\beta(\lambda)$. It holds \new{almost surely}
	\begin{align*}
		β(λ) 
		&≤ \noprkhs*{(\ecovM+λI)^{-1/2}(\ecovM-\scv)(\ecovM+λI)^{-1/2}} \\
	    &= \noprkhs*{(\ecovM+λI)^{-1/2}\ftn n^{-1/2}(I-nH)n^{-1/2}\ftn^*(\ecovM+λI)^{-1/2}}. 
	\end{align*}

	Consider the SVD $\ftn=UΣVᵀ$ with $\new{R} \de \rk(\ftn)$, $U:ℝ^{\new{R}}→\rkhs$, $V \in ℝ^{n×\new{R}}$, \new{$U^*U=I$ and $VᵀV=I$.}
	\new{By definition of $\ecovM=n^{-1}\ftn\ftn^*$}, it holds \new{almost surely that}
	\begin{align*}
		(\ecovM+λI)^{-1/2}\ftn n^{-1/2} = (\ftn\ftn^*+nλI)^{-1/2}\ftn  = (UΣ²U^*+nλI)^{-1/2}UΣVᵀ.
	\end{align*}
	\new{Furthermore, using the fact that $I = P_{\rg(U)^⟂} + P_{\rg(U)}  = P_{\rg(U)^⟂}  + U U^*$ and properties of projection matrices on orthogonal spaces, we have} 
	\begin{align*}
		(\ecovM+λI)^{-1/2}\ftn n^{-1/2}
	    &= (U(Σ²+λnI)U^*+λnP_{\rg(U)^⟂})^{-1/2}UΣVᵀ \\
	    &= (U(Σ²+λnI)^{-1/2}U^*+(λn)^{-1/2}P_{\rg(U)^⟂})UΣVᵀ \\
	    &= U  Σ(Σ²+nλI)^{-1/2} Vᵀ =: UA,
	\end{align*}
	\new{where we defined $A \coloneqq Σ(Σ²+nλI)^{-1/2} Vᵀ ∊ ℝ^{\new{R}×n}$.}
	
	Moreover, using the fact that $‖CBC^*‖=‖(C^*C)^{1/2}B(C^*C)^{1/2}‖$ for any bounded linear operators $B,C$, and $U^*U=I$ we obtain that, almost surely,
	\begin{align*}
		β(λ) 
	    &≤ \noprkhs*{U A (I-nH)A^T U^* } \\
	    &= \noprkhs*{A (I-nH)Aᵀ } \\
	    &= \noprkhs*{AAᵀ - \tfrac{1}{\nf} \sum_{j∊\cI} \tfrac{1}{\pals{j}}a_ja_jᵀ },
	\end{align*}
	where $a_i$ denotes the $i$-th column of $A$. Observe that, for any index $i$, the norm of $a_i$  is equal to the KRR leverage score of observation $i$. Indeed, \new{using standard algebraic operations},
	\begin{align*}
		‖a_i‖²
		&= ‖(Σ²+nλI)^{-1/2} Σ Vᵀe_i‖²\\
		&= e_iᵀ VΣ²(Σ² +λnI)^{-1}Vᵀ e_i \\
		&= e_iᵀ VΣ²Vᵀ(VΣ²Vᵀ +λnI)^{-1} e_i \\
		&= e_iᵀ K (K +λnI)^{-1} e_i = \tls{i},\quad i∊\irange{n}.
	\end{align*}

	\proofparagraph{Step 3: Probabilistic bounds.}
	When $λ≥λ_0$, from the definition of the approximate kernel ridge leverage scores it holds with probability $1-δ/3$	
	\begin{align}
		\pals{i}
		= \frac{\als{i}}{ \sum_{i=1}^n \als{i} }
		&≥ \frac{1}{z²} \frac{\tls{i}}{\sum_{i=1}^n \tls{i}} 
		 = \frac{1}{z²} \frac{‖a_i‖²}{\Tr(AAᵀ)},\quad i∊\irange{n}.
	\end{align}
	Applying \new{\Cref{r:prop9_rudi}} (which is \pcite[Theorem 2]{alaoui2015FastRandomizedKernel} as restated in \pcite[Proposition 9]{rudi2015LessMoreNystrom}), we get with probability $1-\delta/3$	
	\begin{align}
		β(λ)
		&≤ \frac{2 ‖A‖² \log(\tfrac{6n}{δ})}{3\nf} + \sqrt{\frac{ 2‖A‖²z² \Tr(AAᵀ)\log(\tfrac{6n}{δ}) }{\nf} }.
	\end{align}
	It holds $‖A‖²≤‖(Σ²+nλI)^{-1/2}Σ‖²≤1$, and we denote $\Tr(AAᵀ)=\Tr(\ecovM(\ecovM+λI)^{-1})=:\emdeff$. 
	By \Cref{r:joint_effdim}, assuming $\frac{16\supk\log(6/δ)}{n}≤λ≤‖\covM‖$ it holds
	$\emdeff≤2.1\mdeff$ with probability $1-δ/3$. 
	Taking a union bound  
	we get with probability $1-δ$ that
	\begin{align*}
		β(λ)
		&≤ \frac{2 \log(\tfrac{6n}{δ})}{3\nf} + \sqrt{\frac{ 4.2 z² \mdeff\log(\tfrac{6n}{δ}) }{\nf} }
	\end{align*}

	\proofparagraph{Step 4: Simplification.}
	We now derive a condition on $x²\de \log(\tfrac{6n}{δ})/\nf$ in order to ensure $β(λ) ≤ 2/3<1$: 
	by writing the r.h.s as a polynomial in $x$, this holds when $(1/2)x²+(3/4)\sqrt{4.2z²\mdeff}\, x-1/2≤0$, which holds in particular when $0≤x≤r$ where $r$ is the largest root of the second-order polynomial.
	Denoting $b = (3/4)\sqrt{4.2z²\mdeff}$, this root satisfies 
	\begin{align*} 
		r &\de -b+\sqrt{b²+1}=\frac{(\sqrt{b²+1}-\sqrt{b²})(\sqrt{b²+1}+\sqrt{b²})}{(\sqrt{b²+1}+\sqrt{b²})}
		\\& ≥ 1/(2\sqrt{b²+1})
			≥1/(2\sqrt{2.4 z²\mdeff+1}).
	\end{align*}
	Consequently, we have $β(λ) ≤ 2/3$ provided that
	\begin{align*}
		\nf &≥ 4 (1 + 2.4 z²\mdeff) \log(\tfrac{6n}{δ}).
	\end{align*}

	Under this assumption, we thus get by \Cref{e:bypropseven} with probability $1-δ$
	\begin{align*}
		\noprkhs{\Pmo (\ecovM+λI)^{1/2}}²
		&≤ \frac{λ}{1-β(λ)} 
		 ≤ 3λ, 
	\end{align*}
	which concludes the proof.
\end{tproofof*}

\begin{tlemma}{Empirical Two-Sample Effective Dimension}{joint_effdim}
	Let $\delta \in (0, 1)$ and $\lambda > 0$.
	Define the effective dimension and its empirical counterpart as $\mdeff\de\Tr(\covM(\covM+λI)^{-1})$ and $\emdeff\de\Tr(\ecovM(\ecovM+λI)^{-1})$.
	If 
	\begin{align*}
		\frac{16\supk\log(2/δ)}{n}≤λ≤\noprkhs{\covM},
	\end{align*}
	 then, with probability at least $1-\delta$, it holds that
	\begin{align*}
		|\emdeff-\mdeff| \leq 1.1 \mdeff.
	\end{align*}
\end{tlemma}
\begin{tproofof*}{r:joint_effdim}
	In the following, we use the notations $\recovM\de \ecovM+λI$ and $\rcovM\de \covM+λI$. 
	
	\paragraph{Step 1: Deterministic trace inequalities.}
	We aim to show
	\begin{align*}
		|\emdeff-\mdeff| ≤ 2|\Tr(\rcovM^{-1}(\covM - \ecovM)|.
	\end{align*}
	
	Indeed, note that
	\begin{align*}
		|\emdeff-\mdeff|
		&= |\Tr(\recovM^{-1}\ecovM)-\Tr(\rcovM^{-1}\covM)| 		\\
		&≤ |\Tr((\recovM^{-1}-\rcovM^{-1})\ecovM)|+|\Tr(\rcovM^{-1}(\ecovM-\covM))| 		\\
		&= |\Tr(\rcovM^{-1}(\rcovM - \recovM)\recovM^{-1} \ecovM)|+|\Tr(\rcovM^{-1}(\ecovM-\covM))| \\
		&= |\Tr(\rcovM^{-1}(\covM - \ecovM)\recovM^{-1} \ecovM)|+|\Tr(\rcovM^{-1}(\ecovM-\covM))|.
	\end{align*}
	\new{We stress that $\ecovM$ is Hilbert-Schmidt as a finite-rank operator, and $\covM$ is also Hilbert-Schmidt as a trace-class operator, 
	and both $\rcovM^{-1}$ and $\recovM^{-1}$ are bounded operator thanks to the regularization.
	We recall that the product of a Hilbert-Schmidt operator and a bounded operator forms a Hilbert-Schmidt operator, and the sum of two Hilbert-Schmidt operators is also a Hilbert-Schmidt operator, 
	hence both $\rcovM^{-1}(\covM - \ecovM)$ and $\recovM^{-1} \ecovM$ are Hilbert-Schmidt operators.} We can thus use von Neumann's trace inequality \pcite[Theorem 3.1]{carlsson2021NeumannsTraceInequality} and Hölder inequality to obtain 
	\begin{align*}
		|\emdeff-\mdeff| 
		&≤ |\Tr(\rcovM^{-1}(\covM - \ecovM)|\, (\noprkhs{\recovM^{-1} \ecovM}+1) \\
		&≤ 2|\Tr(\rcovM^{-1}(\covM - \ecovM)|.
	\end{align*}

\paragraph{Step 2: Bernstein concentration.}
	
	We now reformulate the problem of controlling the upper bound as one of bounding the deviation of a sum of independent real-valued random variables from their expectation.
	To that end, we introduce the real-valued random variables
	\begin{align*}
		ξ_i \coloneqq \new{\ip{\fmap{W_i}, \rcovM^{-1} \fmap{W_i}}}, \quad i∊\irange{n}.
	\end{align*}
	Using the convention $D(i)=P$ when $i∊\irange{\nX}$ and $D(i) = Q$ when $i∊\irange[\nX+1]{n}$,  
 	it holds $\E[ξ_i]=\Tr(\rcovM^{-1} C_{D(i)})$. In particular, using these notations along with the definitions of $\covM$ and $\ecovM$, \new{and the fact that for any $a,b∊\rkhs$ it holds $\Tr(a\kron b)=\iprkhs{a,b}$,}  we can rewrite the upper bound from the first step as 
	\begin{align*}
		|\emdeff-\mdeff| 
		&≤ \frac{2}{n}|\Tr(\rcovM^{-1}(\nX \ecovX + \nY \ecovY - \nX \covP - \nY \covQ))| \\
		&= \frac{2}{n}|\nX \Tr(\rcovM^{-1}(\ecovX-\covP)) + \nY  \Tr(\rcovM^{-1}(\ecovY-\covQ))|\\
		&= \frac{2}{n}\absv*{ \Tr\prt*{\rcovM^{-1} \sum_{i=1}^{\nX} (\fmap{X_i}\kron\fmap{X_i} - \covP)} 
		  + \Tr\prt*{\rcovM^{-1} \sum_{i=1}^{\nY} (\fmap{Y_i}\kron\fmap{Y_i} - \covQ)}} \\
		&= \frac{2}{n} \absv*{ \sum_{i=1}^n ξ_i -  \E\brk*{\sum_{i=1}^n ξ_i}}.
	\end{align*}
	Let us check the conditions to apply Bernstein's inequality.
	First, note that for any  $i∊\irange{n}$,
	\begin{align*}
		|ξ_i|≤‖\fmap{W_i}‖²‖\rcovM^{-1}‖≤\supk/λ.
	\end{align*}
Second, we have
	\begin{align*}
		\sum_{i=1}^n \E[ξ_i²]
		&= \sum_{i=1}^n \E[\Tr(\iprkhs{\fmap{W_i}, \rcovM^{-1} \fmap{W_i}}\iprkhs{\fmap{W_i}, \rcovM^{-1} \fmap{W_i}})] \\
		&≤ \sum_{i=1}^n \E\brk*{\Tr\prt*{ \rcovM^{-1} (\fmap{W_i}\kron \fmap{W_i})}} \cdot \esssup \absv*{\iprkhs{\fmap{W_i}, \rcovM^{-1} \fmap{W_i}}} \\
		&≤ \supk λ^{-1} \Tr(\rcovM^{-1} (\nX \covP + \nY \covQ)) \\
		&= n\supk λ^{-1}\, \mdeff.
	\end{align*}
	\new{The random variables $(ξ_i)_{1≤i≤n}$ are moreover independent because the data $(\XY_i)_{1≤i≤n}$ is assumed to be drawn \tiid{}, and thus satisfy the assumptions to apply} \Cref{r:real_bernstein_var}: it holds with probability at least $1-δ$ that
	\begin{align*}
		|\emdeff-\mdeff| 
		&≤ 2\prt*{ \sqrt{\frac{2\supk \mdeff\log(2/δ)}{λn}}+\frac{\supk\log(2/δ)}{3λn} } \\
		&= \prt*{ \sqrt{\frac{8q²}{\mdeff }}+\frac{2q²}{3\mdeff} }\,  \mdeff
		\quad\text{where}\quad
		q\de \prt*{\frac{\supk\log(2/δ)}{λn}}^{1/2}.
	\end{align*}

	\paragraph{Step 3: Simplification.}
	Since $λ≤‖\covM‖$, it follows that
	\begin{align*}
		\mdeff ≥ ‖\covM\rcovM^{-1}‖ = ‖\covM‖/(‖\covM‖+λ) ≥ 1/2.
	\end{align*}	
Substituting into the previous bound, we obtain
	\begin{align*}
		|\emdeff-\mdeff| 
		&≤ \prt*{ 4q + \frac{4}{3}q² }\,  \mdeff.
	\end{align*}
	For any $c>0$, the inequality $(4/3)q²+4q≤c$ holds in particular for all $0≤q≤r$ where $r\de (-6+\sqrt{36+12c})/4$ is the largest root of the polynomial $2q²+6q-(3/2)c$.
	Setting $c=1.1$, we get $r≈0.254≥1/4$, so the inequality is satisfied whenever $0≤q²≤1/16$. Therefore, under this condition,
	\begin{align*}
		\P[|\emdeff-\mdeff| \leq 1.1 \mdeff] \geq 1 - δ.
	\end{align*}
\end{tproofof*}

\subsection{Proof of \Cref{r:bound_MMD_nys_new}}
\label{s:proof_bound_MMD_nys_new}

\begin{tproofof*}{r:bound_MMD_nys_new}
We consider the setting of the lemma. By the inverse triangle inequality, we have
\begin{align*}
|\MMD(P,Q)-\ts| 
&= \absv*{ \nrkhs{\kme{P}-\kme{Q}}-\nrkhs{\Pm \kmeX - \Pm \kmeY} } \\
&≤ \nrkhs{\kme{P}-\kme{Q}-\Pm \kmeX + \Pm \kmeY}.
\end{align*}
Applying the standard triangle inequality multiple times then yields
	\begin{align}
		|\MMD(P,Q)-\ts| 
		&≤ \nrkhs{\kme{P}-\kmeX}+\nrkhs{\kme{Q}-\kmeY}+\nrkhs{\Pmo (\kmeX - \kmeY)}. 
			\label{e:mmd_decomposition}
	\end{align}

\paragraph{Step 1: Concentration of the empirical kernel mean embeddings.} 
To control the first two terms in \Cref{e:mmd_decomposition}, we apply \Cref{r:concentration_iid_hilbert} twice—once to the dataset $\X$ and once to the dataset $\Y$—each time with confidence level $\delta/4$. This result is a corollary of \cite[Theorem 3.5]{pinelis1994optimum}; see, for instance, \tcite[\new{Appendix G}]{chatalic2025EfficientNumericalIntegration} for a detailed derivation.
Given that $\sup_i | ϕ(X_i) - \mu_P | \leq 2\supfmap$, it follows from \Cref{r:concentration_iid_hilbert} that, with probability at least $1 - \delta/4$,
\[
\left\| \frac{1}{\nX} \sum_{i=1}^{\nX} \fmap(X_i) - \mu_P \right\|
\leq \frac{2\supfmap \sqrt{2\log(8/\delta)}}{\sqrt{\nX}}.
\]
An analogous bound holds with probability at least $1 - \delta/4$ for the sample $Y$:
\[
\left\| \frac{1}{\nY} \sum_{j=1}^{\nY} \fmap(Y_j) - \mu_Q \right\|
\leq \frac{2\supfmap \sqrt{2\log(8/\delta)}}{\sqrt{\nY}}.
\]

\paragraph{Step 2.1: Impact of the random projection (deterministic inequalities).} 
In order to bound the third term from \eqref{e:mmd_decomposition} we introduce the linear operator
\begin{align*}
	Φ_W\de[\fmap{X_1},…,\fmap{X_{\nX}},\fmap{Y_1},…,\fmap{Y_{\nY}}]:ℝ^n→\rkhs
\end{align*}
and the weight vector \[
w \coloneqq \left[ \tfrac{1}{\nX}, \ldots, \tfrac{1}{\nX},\, -\tfrac{1}{\nY}, \ldots, -\tfrac{1}{\nY} \right] \in \mathbb{R}^{n},
\]
where the first $\nX$ entries are equal to $1/\nX$ and the last $\nY$ entries are equal to $-1/\nY$ so that
\begin{align}
	\nrkhs{\Pmo (\kmeX - \kmeY)} 
	= \nrkhs*{\Pmo Φ_W w} ≤ \n*{\Pmo Φ_W / \sqrt{n} } · \n*{ \sqrt{n}\,  w }. 
	\label{e:ub_term3}
\end{align}

Using the polar decomposition, the linear operator $\Phi_W / \sqrt{n}$ can be written as
\begin{align*}
\Phi_W/\sqrt{n} = \ecovM^{1/2} U,
\end{align*}
for some partial isometry $U$ from $\mathbb{R}^n$ to $\rkhs$. In particular, $\|U\| \leq 1$, and thus 
\begin{align*}
	\n*{\Pmo Φ_W / \sqrt{n} } \leq \noprkhs*{\Pmo (\ecovM)^{1/2}}.
\end{align*}
Let $\lambda>0$. Since $\ecovM \preceq \ecovM + \lambda I$, we have that
\begin{align*}
\noprkhs*{\Pmo (\ecovM)^{1/2}} 
≤ \noprkhs*{\Pmo (\ecovM+λI)^{1/2}}.
\end{align*}

\paragraph{Step 2.2: Impact of the random projection (concentration).} 
Combining these bounds and applying \Cref{r:bound_proj_halfcov_new} with confidence level $δ/2$, we obtain that, with probability at least $1-δ/2$, 
	\begin{align}
		\n*{\Pmo Φ_W / \sqrt{n} } ≤ \sqrt{3λ}, 
		\label{e:ub_lambda}
	\end{align}
	provided the $(\als{i})_{1≤i≤n}$ are $(z,λ_0,δ/6)$-AKRLS and 
	\begin{align}
		\nf &≥ 4 (1 + 2.4 z²\mdeff) \log(\tfrac{12n}{δ}) 
			\label{e:cond1}\\
		λ &≥ λ_0 
			\label{e:cond2} \\
		\frac{16\supk\log(4/δ)}{n} &≤ λ ≤ \noprkhs{\covM}
			\label{e:cond3}
	\end{align}

	We pick
	\begin{align}
		λ=\frac{16\supk \log(4/δ)}{n},
		\label{e:choice_lambda}
	\end{align} 
	so that \cref{e:cond2,e:cond3} are verified under the setting of the lemma,
	and it remains to show that \Cref{e:cond1} hold as well.
	Under \Cref{a:poly_decay_mean}, it holds $\mdeff ≤ c_γ λ^{-γ}$ with $c_γ\de a_γ/(1-γ)$ when $γ<1$ \pcite[Prop. 3]{caponnetto_optimal_2007} and $c_γ\de \supk$ when $γ=1$.
	Hence the r.h.s in \Cref{e:cond1} can be upper-bounded as
	\begin{align*}
		4 (1 + 2.4 z² \mdeff) \log(\tfrac{12n}{δ}) 
		&≤ 4 \prt*{1 + 2.4 z² c_γ \frac{ n^γ }{\prt*{16\supk \log(4/δ)}^{γ}}} \log(\tfrac{12n}{δ}),  
	\end{align*}
	and \Cref{e:cond1} is thus satisfied under the setting of the lemma.

	\paragraph{Step 3: Simplification.} 
	As a consequence, plugging \Cref{e:choice_lambda} in \Cref{e:ub_lambda} and \Cref{e:ub_term3}, it holds with probability at least $1-δ/2$ that
	\begin{align*}
		\nrkhs*{\Pmo (\kmeX - \kmeY)} 
		&≤ \sqrt{3×16\supk \log(4/δ)/n} \sqrt{n} \sqrt{\frac{1}{\nX} +\frac{1}{\nY} } \\
		&≤ 4\supfmap \sqrt{3\log(4/δ)} \prt*{\frac{1}{\sqrt{\nX}} +\frac{1}{\sqrt{\nY}}}.
	\end{align*}

	Putting everything together and 
	using a union bound, we obtain that, with probability at least $1-δ$,
	\begin{align*}
		|\MMD(P,Q)-\ts| 
		&≤ \prt*{ 2\supfmap \sqrt{2\log(8/δ)}
		+ 4\supfmap \sqrt{3\log(4/δ)} }\prt*{\frac{1}{\sqrt{\nX}} +\frac{1}{\sqrt{\nY}} }
		\\&≤ 10\supfmap \sqrt{\log(8/δ)}\prt*{\frac{1}{\sqrt{\nX}} +\frac{1}{\sqrt{\nY}}}.
	\end{align*}
\end{tproofof*}

\section{Controlling the power}

\subsection{Proof of \Cref{r:bound_threshold}}

\begin{tproofof*}{r:bound_threshold}{}
	Define the events 
	$\cA=\cb{\ts≤\thr}$ and
	\[\cB_{\beta}=\cb{\MMD(P,Q)≥ \mathcal{E}_{MMD}(n_X, n_Y, \beta/2) +\thr}.\] 
	\new{We recall that we assume} $\P[\cB_{\beta}]>1-β/2$. We will show that $\P[\cA]≤β$.
	
	By the law of total probability, it holds
	\begin{align*}
		\P[\cA]
		&= \P[\cA \cap \cB_\beta]
		 + \P[\cA|\cB_\beta^c] \P[\cB_\beta^c] ≤ \P[\cA \cap \cB_\beta] 
		 + \P[\cB_\beta^c].
		 \label{e:first_ineq_cA}
	\end{align*}
	By assumption, the first term in the above upper bound can be bounded as 
	\begin{align*}
		\P[\cA \cap \cB_\beta] 
		&= \P[\ts≤\thr\, , \, \thr≤\MMD(P,Q)-\eMMD[\nicefrac{\beta}{2}]] \\
		&≤ \P[\ts≤\MMD(P,Q)-\eMMD[\nicefrac{\beta}{2}]] \\
		&≤ \P[|\MMD(P,Q)-\ts|≥\eMMD[\nicefrac{\beta}{2}]]\\
		&\leq \beta/2 \enspace.
	\end{align*}
	We get the desired result by combining the last inequality and the fact that $\P[\cB_\beta^c]≤\beta/2$.
\end{tproofof*}

\subsection{Bounds on the quantiles}
\label{s:decomposition_U_R}

In order to prove \Cref{r:bound_TS_quantile_rough} and  \Cref{r:bound_quantile_rough} below, we first show how the squared test statistic can we decomposed as the sum of a $U$-statistic and a remainder, and provide a bound on the quantile of this $U$-statistic.
The (randomly) permuted squared test statistic can be written 
\begin{align*}
	\pts² 
	&\de \n*{\tfrac{1}{\nX} \sum_{i=1}^{\nX} \tfmap(\XY_{σ(i)}) - \tfrac{1}{\nY} \sum_{j=1}^{\nY} \tfmap(\XY_{σ(\nX+j)})}²\\
	&\de \n*{\tfrac{1}{\nY\nX} \sum_{i=1}^{\nX}\sum_{j=1}^{\nY} \prt*{\tfmap(\XY_{σ(i)}) - \tfmap(\XY_{σ(\nX+j)}) } }²\\
	&= \tfrac{1}{\nX²\nY²} \sum_{i=1}^{\nX}\sum_{j=1}^{\nY}\sum_{i'=1}^{\nX}\sum_{\new{j'}=1}^{\nY}  
		h^σ(i,i';j,j') 
\end{align*}
with $ h^σ(i,i';j,j') \de  \tk{\XY_{σ(i)},\XY_{σ(i')}} -  \tk{\XY_{σ(i)}, \XY_{σ(\nY+j')}} - \tk{\XY_{σ(\nY+j)},\XY_{σ(i')}} + \tk{\XY_{σ(\nY+j)}, \XY_{σ(\nY+j')}}$ where $\tk(x,y)\de\ip{\tfmap{x},\tfmap{y}}$.

It can be decomposed as the sum of a (weighted) U-statistic and a remainder term,
\begin{align}
	\pts² 
	&= \tfrac{(\nX-1)(\nY-1)}{\nX\nY} \pstsU + \pstsR 
		\label{e:ts_decomposition},
\end{align}
with
\begin{align}
	\pstsU 
	&\de 
		\tfrac{1}{\nX(\nX-1)\nY(\nY-1)} \sum_{1≤i≠i'≤\nX} \sum_{1≤j≠j'≤\nY}  
		h^σ(i,i';j,j'),
		\label{e:def_U}
\end{align}
and
\begin{align}
	\pstsR
	&= \tfrac{1}{\nX²\nY²} 
		\prt*{\sum_{i=1}^{\nX}\sum_{1≤j≠j'≤\nY}  h^σ(i,i;j,j')
		+ \sum_{1≤i≠i'≤\nX}\sum_{j=1}^{\nY} h^σ(i,i';j,j) 
		+ \sum_{i=1}^{\nX}\sum_{j=1}^{\nY}  h^σ(i,i;j,j)
		}.
	\label{e:def_R}
\end{align}
We control  the $U$-statistic with high probability using a result from \tcite{kim2022MinimaxOptimalityPermutationa} while we obtain an upper-bound for $\pstsR$ using the boundedness assumption.

\begin{tlemma}{Bound on the quantile of $\pstsU$ given $\X,\Y$}{bound_U_quantile_rough}
Let $0<α≤e^{-1}$. Then there is a universal constant $C'$ such that the $U$-statistic \eqref{e:def_U} associated to the Nyström kernel approximation, and depending on the permutation $σ$ uniformly drawn from $\new{S_n}$, satisfies
\begin{align*}
	\P*[\pstsU ≤ C' \tfrac{\supk}{\sqrt{\nX(\nX-1)}} \log(1/α) \given \X,\Y] 
		&≥ 1-α.
\end{align*}
That is, the r.h.s. of the above is an upper bound for the quantile $\pstsUq$.
\end{tlemma}
\begin{tproofof*}{r:bound_U_quantile_rough}
Let
\begin{align*}
	Σ_{\nX,\nY}²
		&\de \tfrac{1}{\nX²(\nX-1)²} 
			\sup_{σ∊\new{S_n}}
		\sum_{1≤i≠i'≤\nX} \ip{\tfmap{\XY_{σ(i)}},\tfmap{\XY_{σ(i')}}}².
\end{align*}
Given that $\Pm$ is a projector, it holds for any $x,y$ that $\tk(x,y)²=\ip{\Pm\fmap{x},\Pm\fmap{y}}²≤‖\fmap{x}‖²‖\fmap{y}‖²$, and thus
\begin{align*}
	 Σ_{\nX,\nY}²
		&≤ \tfrac{\supk²}{\nX(\nX-1)}.
\end{align*}
By \tcite[Th. 6.1 and \new{(22)}]{kim2022MinimaxOptimalityPermutationa}, it holds conditionally on $\X,\Y$, with probability at least $1-α$ that
\begin{align*}
	\pstsU 
	&≤ \max\prt*{ \sqrt{C^{-1} Σ_{\nX,\nY}² \log(1/α) }, C^{-1} Σ_{\nX,\nY} \log(1/α)} \\
	&≤ C' Σ_{\nX,\nY} \log(1/α),
\end{align*}
where $C$ refers to the absolute constant of \tcite[Th. 6.1]{kim2022MinimaxOptimalityPermutationa} and $C'$ is another absolute constant.
\end{tproofof*}

We can now prove \Cref{r:bound_TS_quantile_rough}.

\begin{tproofof*}{r:bound_TS_quantile_rough}
Starting from \Cref{e:ts_decomposition}, we have
\begin{align*}
	\pts^2
	&= \tfrac{(\nX-1)(\nY-1)}{\nX\nY} \pstsU + \pstsR.
\end{align*}
Since $\Pm$ is a projector,  $\tk(x,y)=\ip{\Pm\fmap{x},\Pm\fmap{y}}≤‖\fmap{x}‖‖\fmap{y}‖$. It follows that, almost surely,
\begin{align*}
	|\pstsR|
	&≤ \tfrac{(\nY+\nX-1)}{\nX\nY} 4\supk
	 ≤ (\tfrac{1}{\nX} +\tfrac{1}{\nY}) 4\supk.
\end{align*}
Moreover, by \Cref{r:bound_U_quantile_rough}, conditionally on $\X,\Y$, it holds with probability at least $1-α$  that
\begin{align*}
	\pstsU ≤ C' \tfrac{\supk}{\sqrt{\nX(\nX-1)}} \log(1/α).
\end{align*}
Combining everything together, it holds with probability at least $1-\alpha$ that
\begin{align*}
	\pts^2
	&≤ C' \tfrac{\supk}{\nX} \log(1/α) 
		+ (\tfrac{1}{\nX} +\tfrac{1}{\nY}) 4\supk.
\end{align*}
\end{tproofof*}

We now prove our high-probability bound on the threshold (\Cref{r:bound_quantile_rough}).

\begin{tproofof*}{r:bound_quantile_rough}
	Using \pcite[Lemma 6]{fixed_domingo-enrich2023CompressThenTest}, it holds
	\begin{align*}
		\P*[\thr ≤ q_{1-α_1}(\X,\Y) \given \X,\Y] &> 1 - \beta/2 
	\end{align*}
	where $α_1\de \prt*{\frac{ \beta/2}{\binom{\np}{\lfloor α(\np+1) \rfloor}}}^{1/\lfloor α(\np+1) \rfloor}$.
	\new{Let $\tq$ denote the $(1-α)$-quantile (with respect to the random variable $σ$) for the test statistic $\pts$. }
	By \Cref{r:bound_TS_quantile_rough}, provided $α₁≤e^{-1}$, it holds (almost surely for $\X,\Y$)
	\begin{align*}
		\tq[α₁] 
		&≤ \supfmap \sqrt{C' \tfrac{1}{\nX} \log(1/α₁) + (\tfrac{4}{\nX} +\tfrac{4}{\nY})}\\
		&≤ \sqrt{C'}\supfmap \tfrac{1 }{\sqrt{\nX}} \sqrt{\log(1/α₁)} + (\tfrac{1}{\sqrt{\nX}} +\tfrac{1}{\sqrt{\nY}}) 2\supfmap\enspace.
	\end{align*}

	We will now upper bound $\log(1/\alpha_1)$ to make our bound more explicit.
	Denoting $c_α=\lfloor α(\np+1) \rfloor$, it holds $c_α≥1$ by assumption. Using the inequalities $\binom{n}{k}≤\prt*{\frac{en}{k}}^k$ and $\lfloor x \rfloor \geq x/2$, we obtain
	\begin{align*}
		\log\prt*{\binom{\np}{c_α}^{1/c_α}} 
		&≤ \log\prt*{\frac{e \np}{c_α}} ≤ \log\prt*{\frac{2 e \np}{α(\np+1)}} \leq \log\prt*{\frac{2e}{α}}\enspace.
	\end{align*}
	The population \new{(with respect to the random variable $σ$)} quantile can then be upper bounded as
	\begin{align*}
		\tq[α₁] 
		&≤ \sqrt{C'}\supfmap \tfrac{1}{\sqrt{\nX}} \sqrt{\log\prt*{\frac{2e}{α (\beta/2)^{1/c_α}}}} + (\tfrac{1}{\sqrt{\nX}} +\tfrac{1}{\sqrt{\nY}}) 2\supfmap
		\enspace.
	\end{align*}
	We finish the proof by noting that the assumption $α_1≤1/e$ is not restrictive. Indeed
	\begin{align*}
		α_1 
		&= \frac{(β/2)^{1/c_α}}{\binom{\np}{c_α}^{1/c_α}} 
		 ≤ \frac{1}{(\frac{\np}{c_α})}
		 ≤ \frac{α(\np+1)}{\np} 
		 ≤ 2α
	\end{align*}
	which is upper-bounded by $e^{-1}$ under our assumptions.
\end{tproofof*}

\subsection{Proof of \Cref{r:power_MMD} (main result)}
\label{s:proof_main_result}

Let $\tq$ denote the $(1-α)$-quantile (with respect to the random variable $σ$) for the test statistic $\pts$. 

\begin{tproofof*}{r:power_MMD}{}
We apply \Cref{r:bound_MMD_nys_new} with $δ=β/2$. 
Defining 
\begin{align*}
	\eMMD\de 10 \supfmap \sqrt{\log(8/δ)} \left( \frac{1}{\sqrt{\nX}} + \frac{1}{\sqrt{\nY}} \right)
\end{align*}
as in \Cref{r:bound_MMD_nys_new}, it holds under our assumptions
\begin{align*}
	\P[|\MMD(P,Q)-\ts|≥\eMMD[β/2]] &≤ β/2. 
\end{align*}

Consequently, in order to lower bound the power of the test by $1-β$, it is sufficient by \Cref{r:bound_threshold} to prove
\begin{align*}
	\P[\thr≤\MMD(P,Q)-\eMMD[\beta/2]] > 1- \beta/2\enspace.
\end{align*}

By \Cref{r:bound_quantile_rough} (for which the requirement on $α$ is satisfied under our assumptions) it holds 
\begin{align*}
	\P*[\thr ≤ \sqrt{C'}\supfmap \tfrac{1 }{\sqrt{\nX}} \sqrt{\log\prt*{\frac{2e}{α (\beta/2)^{1/c_α}}}} + (\tfrac{1}{\sqrt{\nX}} +\tfrac{1}{\sqrt{\nY}}) 2\supfmap \given \X,\Y] &> 1 - β/2.
\end{align*}

Hence, we obtain the desired power bound provided
\begin{align*}
	\sqrt{C'}\supfmap \tfrac{1}{\sqrt{\nX}} \sqrt{\log\prt*{\frac{2e}{α} \prt*{2/β}^{1/c_α}} } + (\tfrac{1}{\sqrt{\nX}} +\tfrac{1}{\sqrt{\nY}}) 2\supfmap
	&≤ \MMD(P,Q) 
	 - \eMMD[β/2],
\end{align*}
that is, for $c_α\de\lfloor α(\np+1) \rfloor$,
\begin{align*}
	\MMD(P,Q) 
		&≥ 
	\sqrt{C'}\supfmap \tfrac{1 }{\sqrt{\nX}} \sqrt{\log\prt*{\frac{2e}{α} \prt*{\frac{2}{β}}^{1/\lfloor α(\np+1) \rfloor}} } 
	\\&\quad
		+ 2\supfmap\prt{1 + 5  \sqrt{\log(16/β)}}(\tfrac{1}{\sqrt{\nX}} +\tfrac{1}{\sqrt{\nY}}) \new{\,=: (\star)}.
\end{align*}
\new{
This yields the claimed result by noting that the square of the right-hand-side in this last equation can be upper-bounded as follows:
\begin{align*}
	(\star)²
		&≤ 2\prt*{C'\supk \tfrac{1 }{\nX} \log\prt*{\frac{2e}{α} \prt*{\frac{2}{β}}^{1/\lfloor α(\np+1) \rfloor}} + 8\supk\prt{1 + 5  \sqrt{\log\prt*{\frac{16}{β}}}}²(\tfrac{1}{\nX} +\tfrac{1}{\nY})} \\
		&≤ \frac{2\supk}{\nX}\prt*{C' \log\prt*{\frac{2e}{α} \prt*{\frac{2}{β}}^{1/\lfloor α(\np+1) \rfloor}} + 16\prt*{6  \sqrt{\log\prt*{\frac{16}{β}}}}²} \\
		&≤ \frac{2\supk \max(C', 16×36)}{\nX}\prt*{\log\prt*{\frac{2e}{α} \prt*{\frac{2}{β}}^{1/\lfloor α(\np+1) \rfloor}} + \log\prt*{\frac{16}{β}}} 
\end{align*}
where we used $\nX/\nY≤1$, and the fact that $\sqrt{\log(16/β)}>1$ holds given that $β≤1$.
}
\end{tproofof*}

\ifejs
	\end{appendix}%
\else\fi

\end{document}